\PassOptionsToPackage{table}{xcolor}
\documentclass{navercloud}
\usepackage[T1]{fontenc}
\usepackage{times}

\usepackage{amsmath,amsfonts,bm}

\def\eqref#1{equation~\ref{#1}}

\def\1{\bm{1}}

\def\mI{{\bm{I}}}

\DeclareMathAlphabet{\mathsfit}{\encodingdefault}{\sfdefault}{m}{sl}
\SetMathAlphabet{\mathsfit}{bold}{\encodingdefault}{\sfdefault}{bx}{n}

\newcommand{\E}{\mathbb{E}}

\newcommand{\R}{\mathbb{R}}

\usepackage{booktabs}
\usepackage{hyperref}
\usepackage{url}
\usepackage{graphicx}
\usepackage{subcaption}

\usepackage{enumitem}
\usepackage{wrapfig}

\newcommand{\thinhyp}{\kern-0.08em\nobreak-\nobreak\kern-0.08em}

\DeclareRobustCommand{\captitle}[1]{\opthead{#1}}

\title{Let Vision-Language Models Generate Images through Their Representations}

\author[1, 2, *]{JB~Kim}
\author[2]{Byungwoo~Jeon}
\author[2]{Kyungmin~Lee}
\author[2]{Hyeonmo~Kang}
\author[1,2]{Cheonbok~Park}
\author[1]{Shinyoung~Joo}
\author[2\dagger]{Jinwoo~Shin}
\author[1,2\dagger]{Jeonghoon~Kim}

\affiliation{$^1$NAVER Cloud, $^2$Korea Advanced Institute of Science and Technology}
\contribution[*]{Work done during the NAVER Cloud Residency Program}
\contribution[\dagger]{Corresponding authors}

\abstract{
Recent unified multimodal models aim to combine visual understanding and generation within a single model, yet effectively integrating the two capabilities remains challenging. Leveraging pretrained vision-language models offers a way to build on already learned multimodal knowledge, but making this knowledge effectively accessible to visual generation remains nontrivial.
We find that a central challenge lies in the heterogeneous visual representations used for understanding and generation.
To address this challenge, we introduce \textit{Integral}, which converts a pretrained vision-language model into a unified multimodal model by constructing a shared visual representation space across the two capabilities.
\textit{Integral} implements this design through two complementary components. Abstraction Fusion constructs shared visual tokens by aggregating hierarchical representations from the frozen vision encoder, while Block-wise MoT introduces a lightweight generative stream that interacts with the vision-language model only at block boundaries.
Our approach effectively preserves the pretrained understanding capability while extending it to strong visual generation.
Controlled comparisons further demonstrate that sharing visual representations between understanding and generation substantially improves generative learning compared with using a separate visual representation space, despite comparable reconstruction fidelity. These results suggest that representations learned by pretrained vision-language models already provide an effective foundation for visual generation.
}

\metadata[\nanumsquarefont Date]{\today}
\correspondence{\email{jb@thejb.net}, \email{jinwoo@kaist.ac.kr}, \email{jeonghoon.samuel@gmail.com}}
\metadata[\nanumsquarefont Project Page]{\url{https://thejb.ai/integral}}

\begin{document}

\maketitle

\vspace{10pt}
\newlength{\teaserheight}
\setlength{\teaserheight}{0.23\textwidth}
\begin{center}
  \captionsetup{type=figure}
  \begin{minipage}[b]{0.378\textwidth}
    \centering
    \includegraphics[height=\teaserheight]{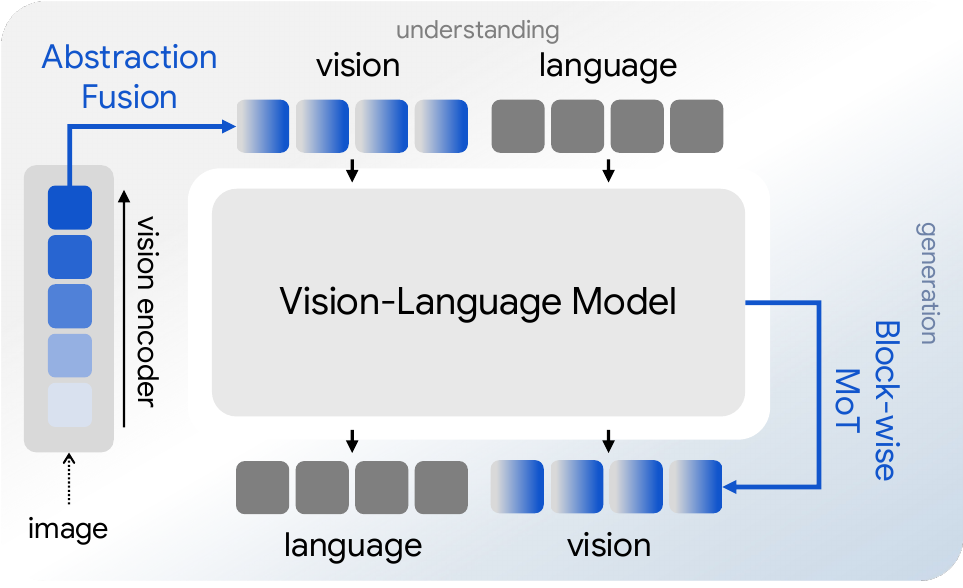}
  \end{minipage}\hfill%
  \begin{minipage}[b]{0.305\textwidth}
    \centering
    \includegraphics[height=\teaserheight]{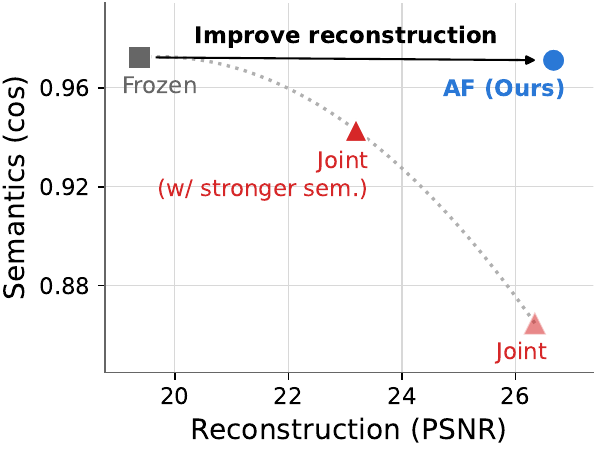}
  \end{minipage}\hfill%
  \begin{minipage}[b]{0.292\textwidth}
    \centering
    \includegraphics[height=\teaserheight]{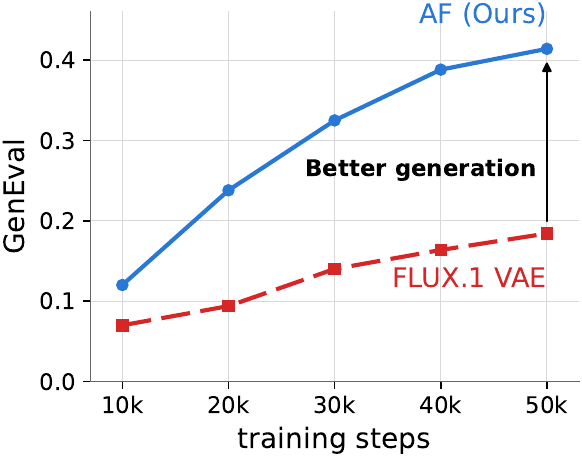}
  \end{minipage}
\caption{\captitle{Overview.}
\textit{Integral} extends pretrained vision-language models (VLMs) with visual generation through shared visual representations.
Abstraction Fusion (AF) enables fine-grained reconstruction without compromising semantic quality, while Block-wise MoT introduces a lightweight generative pathway conditioned on the VLM.
This shared representation facilitates effective generative learning by leveraging pretrained multimodal knowledge.
}

  \label{fig:teaser}
\end{center}

\section{Introduction}

\begin{figure}[t]
\centering
\includegraphics[width=\textwidth]{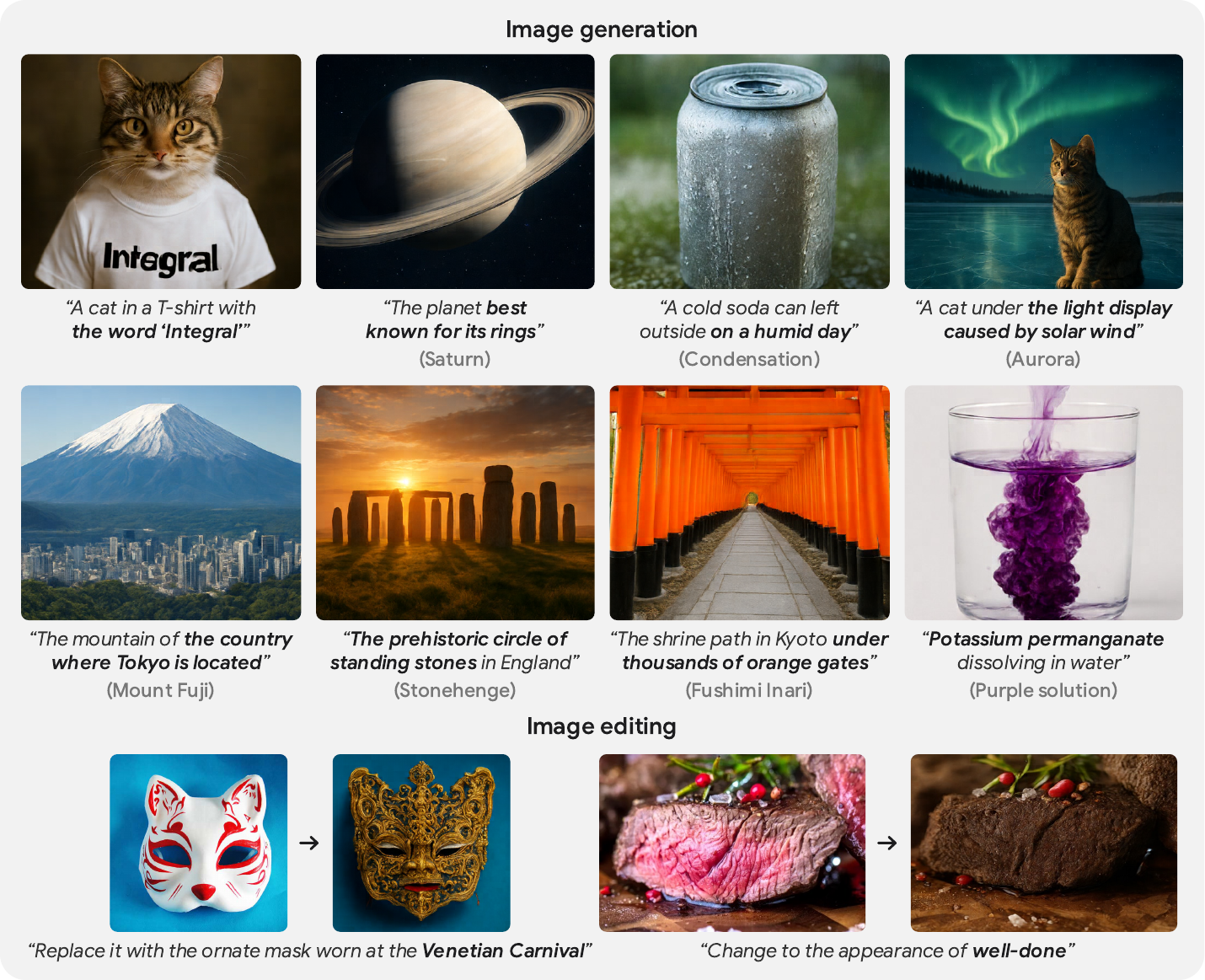}
\vspace{-15pt}
\caption{\captitle{Qualitative examples of generation and editing.} Prompts are abbreviated; bold text marks the part requiring diverse world knowledge, with its meaning in parentheses. Examples include WISE~\citep{niu2026wise} and KRIS-Bench~\citep{wu2025kris} prompts.}
\label{fig:qual}
\vspace{-8pt}
\end{figure}

Unifying multimodal understanding and generation within a single model has received growing attention in multimodal learning~\citep{openai2024gpt4ocard,chameleon2024chameleon,chen2025janus,wang2024emu3}.
Such unification offers the potential for visual generation to leverage knowledge acquired through multimodal understanding~\citep{openai2024gpt4ocard}.
However, jointly optimizing for understanding and generation remains challenging, as the two objectives can interfere with each other, potentially compromising understanding performance~\citep{hao2026unix,ye2026understanding,liu2026symbiotic}.
To mitigate such interference, recent unified multimodal models (UMMs) introduce generation-specific capacity to create a separate pathway for visual generation~\citep{pan2025transfer,shi2026lmfusion,deng2025bagel}.
Such specialization poses a challenge for the generation pathway to effectively leverage knowledge encoded in the understanding pathway.

Prior work has explored architectural designs that connect pretrained vision-language models (VLMs) to generative pathways through learnable queries~\citep{pan2025transfer}, feature conditioning~\citep{wu2025omnigen2}, or shared attention~\citep{wang2026lightfusion,shen2025latteflow}.
While these approaches allow generative pathways to access pretrained VLM representations, they typically model generation in a separate, heterogeneous visual representation space, often defined by reconstruction-oriented VAEs~\citep{esser2024scaling,flux2024,wu2025qwenimagetechnicalreport}.
Recent studies have highlighted the role of sharing visual representations in leveraging pretrained multimodal knowledge for generation~\citep{wu2025harmonizing,kang2026transferability}.
However, directly using VLM representations for generation remains challenging, as their high-level features often lack the fine-grained information required for image reconstruction~\citep{chen2025blip3,zhang2025both}.
Adapting the vision encoder for reconstruction, in turn, can degrade its pretrained semantic capabilities~\citep{wu2025vila,lin2025toklip,yue2026uniflow}.
This raises the question of whether rich representations learned by pretrained VLMs for multimodal understanding can also be leveraged for visual generation without compromising their understanding capabilities.

In this work, we propose \textit{Integral}, a unified multimodal modeling approach that extends a pretrained vision-language model with visual generation in a shared visual representation space while preserving its understanding capabilities. Our analysis suggests that a key bottleneck in adapting pretrained vision encoders for reconstruction while preserving their semantic quality may lie in the retention of low-level features in deeper layers, potentially interfering with progressive semantic abstraction~(Sec.~\ref{sec:bottleneck}).
Based on this observation, we introduce Abstraction Fusion~(AF), which aggregates representations across multiple abstraction levels of the frozen vision encoder into shared visual tokens $z$ for understanding and generation without altering the pretrained hierarchy. To efficiently leverage the understanding pathway for generation in this token space, we further introduce Block-wise Mixture-of-Transformers~(Block-wise MoT), a lightweight generative architecture that conditions on the VLM's internal representations at block boundaries.

\textbf{Contributions.}
We introduce \textit{Integral}, which extends a pretrained vision-language model with strong visual generation using only 0.89B generation-specific parameters while preserving its understanding capability.
\textit{Integral} achieves strong performance across compositional generation, knowledge-intensive generation, and image editing that requires multimodal understanding.
Controlled comparisons show that sharing the visual representation space improves generative learning beyond what can be explained by reconstruction quality alone.

\section{Method}
\label{sec:method}

We introduce \textit{Integral}, a unified multimodal modeling approach that extends a pretrained vision-language model with visual generation through a shared visual token space while preserving its understanding capabilities.
\textit{Integral} consists of two key components.
Abstraction Fusion (AF) constructs visual tokens $z$ that are used for both multimodal understanding and high-fidelity image reconstruction~(Sec.~\ref{sec:method-af}).
Block-wise MoT is a lightweight generative architecture that learns a conditional velocity field over $z$ using VLM representations at block boundaries~(Sec.~\ref{sec:method-mot}).

\subsection{Propagation Bottleneck}
\label{sec:bottleneck}

\begin{figure}[t]
  \centering
  \begin{subfigure}[t]{0.260\textwidth}
    \centering
    \includegraphics[width=\linewidth]{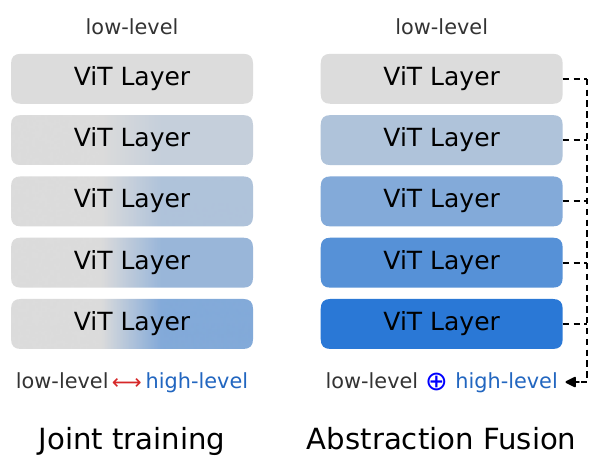}
    \caption{Joint vs. AF (Ours)}
    \label{fig:bottleneck-a}
  \end{subfigure}\hfill%
  \begin{subfigure}[t]{0.259\textwidth}
    \centering
    \includegraphics[width=\linewidth]{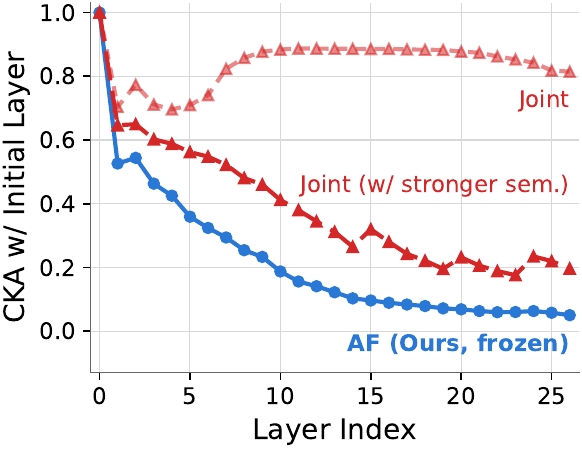}
    \caption{Layer-wise similarity}
    \label{fig:bottleneck-b}
  \end{subfigure}\hfill%
  \begin{subfigure}[t]{0.456\textwidth}
    \centering
    \includegraphics[width=\linewidth]{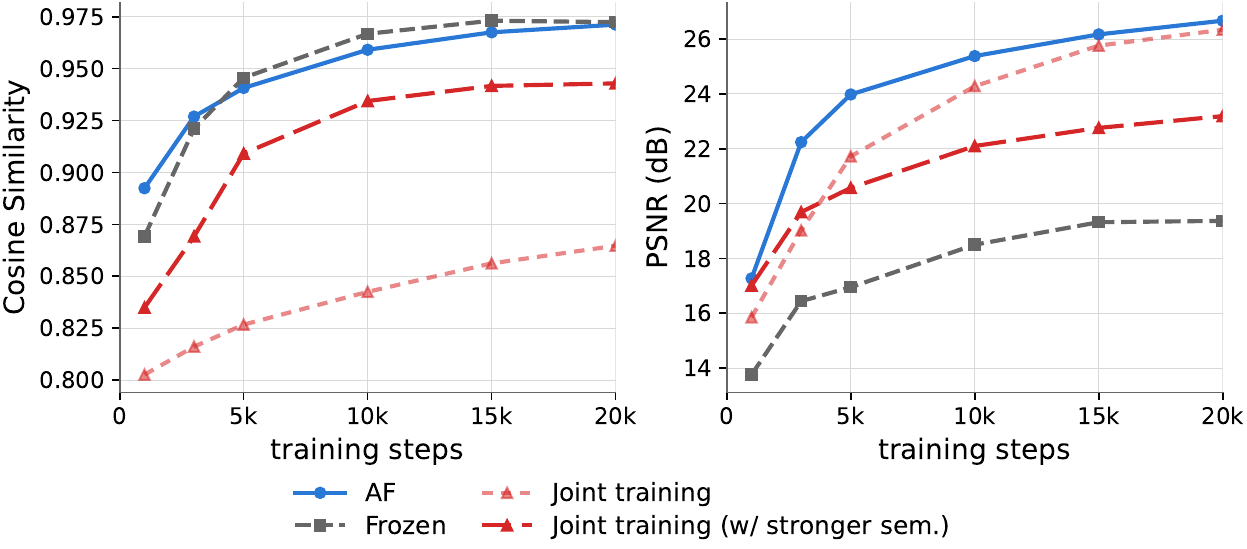}
    \caption{Semantics and reconstruction during training}
    \label{fig:bottleneck-c}
  \end{subfigure}
\caption{
\captitle{Propagation bottleneck and Abstraction Fusion (AF).}
(a) Joint training of the encoder requires low-level visual information to propagate to the final layer, whereas AF directly fuses features across the frozen encoder hierarchy.
(b) Layer-wise CKA analysis shows that joint training increases the similarity between deeper layers and the initial-layer.
(c) Joint training exhibits a trade-off between semantic preservation and reconstruction fidelity during training, whereas AF largely avoids this trade-off.
}
  \label{fig:bottleneck}
  \vspace{-8pt}
\end{figure}

Our goal is to extend the representations learned by a pretrained vision encoder to support image reconstruction while preserving their semantic quality, serving as a shared space in both understanding and generation.
A key challenge is that representations become increasingly abstract across encoder depth, making fine-grained visual details difficult to recover from the final representation~\citep{mahendran2015understanding,li2024cascade}.
Adapting the vision encoder for reconstruction improves reconstruction fidelity but can degrade semantic quality, reflecting the semantics--reconstruction trade-off shown in Fig.~\ref{fig:bottleneck-c} and observed in prior unified representation learning~\citep{wu2025vila,ma2025unitok,yang2026muse,yue2026uniflow}.

Prior work has primarily attributed the semantics--reconstruction trade-off to optimization difficulties or limited representation capacity~\citep{ma2025unitok,yang2026muse,fan2025prism}.
We investigate a complementary explanation based on how information is transformed across the depth of the vision encoder.
Vision encoders progressively transform local visual information into more abstract semantic representations~\citep{yosinski2014transferable,mahendran2015understanding}, while intermediate layers retain fine-grained visual details that are increasingly attenuated in deeper representations~\citep{li2024cascade,zhang2024learning}.
When reconstruction relies only on the final representation, training may encourage deeper layers to preserve information characteristic of earlier representations, potentially limiting progressive semantic abstraction (Fig.~\ref{fig:bottleneck-a}).
We refer to this hypothesis as a \textit{propagation bottleneck}.

To examine this hypothesis, we analyze how joint training alters the representation hierarchy of the vision encoder.
This view anticipates that deeper representations would become more similar to the initial-layer representation under joint training.
We therefore measure the CKA similarity between each encoder layer and the initial-layer representation.
As shown in Fig.~\ref{fig:bottleneck-b}, joint training of the vision encoder substantially increases this similarity in deeper layers relative to the pretrained encoder.
Increasing semantic supervision reduces this similarity, consistent with greater representational differentiation across depth.
Fig.~\ref{fig:bottleneck-c} shows a trade-off between semantic preservation and reconstruction fidelity during training, with stronger semantic supervision degrading reconstruction fidelity.
Together, these results are consistent with our \textit{propagation bottleneck} hypothesis.

These observations suggest avoiding exclusive reliance on the final encoder representation for reconstruction.
This motivates us to propose Abstraction Fusion (AF), which constructs visual tokens by aggregating representations across multiple abstraction levels of the frozen vision encoder without altering its encoder hierarchy~(Fig.~\ref{fig:bottleneck-a}).
AF substantially improves reconstruction fidelity while maintaining semantics close to those of the frozen encoder~(Fig.~\ref{fig:bottleneck-c}).

\subsection{Visual Tokenization with Abstraction Fusion}
\label{sec:method-af}

\begin{figure}[t]
  \centering
  \begin{minipage}[c]{0.46\textwidth}
    \centering
    \includegraphics[width=\linewidth]{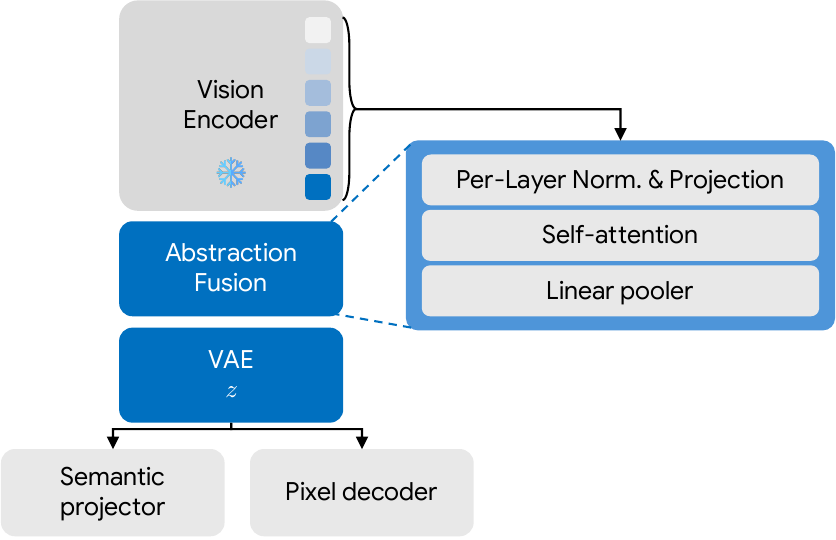}
    \caption{
\captitle{Abstraction Fusion.}
AF aggregates multi-level representations from a frozen vision encoder via per-layer normalization and projection, self-attention, and a linear pooler.
}
    \label{fig:tokenizer}
  \end{minipage}\hfill
  \begin{minipage}[c]{0.5\textwidth}
    \centering
    \captionof{table}{\captitle{Image reconstruction quality.}
    Reconstruction on ImageNet-1k at $256{\times}256$ resolution.}
    \label{tab:reconstruction}
    \setlength{\tabcolsep}{3pt}
\small
\begin{tabular}{lccc}
\toprule
Visual Tokenizer & rFID$\downarrow$ & PSNR$\uparrow$ & SSIM$\uparrow$ \\
\midrule
\multicolumn{4}{l}{\textit{Unified tokenizers}} \\
RAE w/ Dinov2~\citep{zheng2026diffusion} & 0.49 & 18.1 & 0.478 \\
RAE w/ SigLIP2~\citep{zheng2026diffusion} & 0.44 & 21.3 & 0.593 \\
VTP~\citep{yao2026towards} & 0.36 & 25.8 & 0.741 \\
PS-VAE~\citep{zhang2025both} & 0.20 & 28.8 & 0.817 \\
\midrule
\multicolumn{4}{l}{\textit{Reconstruction-only tokenizers}} \\
SD-VAE~\citep{rombach2022high} & 0.90 & 24.2 & -- \\
VA-VAE~\citep{yao2025reconstruction} & 0.28 & 28.0 & 0.790 \\
FLUX.1-VAE~\citep{flux2024} & 0.17 & 31.1 & -- \\
\midrule
\rowcolor{gray!20}
\textbf{Qwen3.5-9B w/ AF (ours)} & \textbf{0.15} & \textbf{32.2} & \textbf{0.942} \\
\bottomrule
\end{tabular}

  \end{minipage}

  \vspace{8pt}
  \captionof{table}{\captitle{Visual understanding w/ AF.}
  Key performances before and after applying AF on Qwen3.5-9B. Avg.\ Retention$_{20}$ averages retention over 20 benchmarks provided in Appendix~\ref{app:understanding}.}
  \label{tab:understanding}
  \vspace{4pt}
  {\setlength{\tabcolsep}{3pt}
\resizebox{\textwidth}{!}{%
\begin{tabular}{lcccccccccc}
\toprule
Model & POPE$\uparrow$ & MME$\uparrow$ & MMB$\uparrow$ & SEED$\uparrow$ & SEED2+$\uparrow$ & MMStar$\uparrow$ & MMMU$\uparrow$ & RWQA$\uparrow$ & OCRBench$\uparrow$ & Avg. Retention$_{20}\uparrow$ \\
\midrule
Qwen3.5-9B & 89.6 & 2400 & 86.3 & 78.9 & 73.1 & 75.9 & 73.2 & 78.2 & 88.1 & 100.0\% \\
\rowcolor{gray!20}
Qwen3.5-9B w/ AF & 89.6 & 2412 & 85.4 & 78.8 & 73.2 & 76.0 & 73.5 & 78.7 & 87.3 & 99.8\% \\
\bottomrule
\end{tabular}}
}
  \vspace{-10pt}
\end{figure}

We introduce Abstraction Fusion (AF), a visual tokenization method that integrates representations
from multiple abstraction levels of a frozen vision encoder
to construct unified visual tokens $z$.

\textbf{Abstraction Fusion.}
Given an image $x$, we extract hierarchical representations
$\{h^{(\ell)}\}_{\ell=0}^{L}$ from the encoder.
AF aggregates these representations into a fused representation
$f$ while preserving the spatial token structure:
\begin{equation}
    f
    =
    \mathrm{AF}\!\left(\{h^{(\ell)}\}_{\ell=0}^{L}\right),
    \qquad
    h^{(\ell)} \in \R^{N \times d_h},
    \quad
    f \in \R^{N \times d_f},
\end{equation}
where $N$ denotes the number of spatial tokens, $d_h$ and $d_f$ the feature dimensions, $L$ the number of encoder layers, and $h^{(0)}$ the patch embedding.

We implement the feature fusion in AF with three key components, as illustrated in Fig.~\ref{fig:tokenizer}.
Per-layer normalization and projection first map representations from different encoder layers into a common feature space.
For each spatial position, we then apply self-attention across the layer dimension to aggregate information from different abstraction levels.
A linear pooler reduces the resulting layer-wise features into a single fused token at each spatial position.
The fused representation is finally compressed through a variational bottleneck into visual tokens $z \in \R^{N \times d_z}$, where $d_z$ denotes the latent dimension.


The resulting visual tokens $z$ serve as a shared representation for multimodal understanding and image generation. For understanding, an MLP projector $g_{\mathrm{sem}}$ maps $z$ to $\hat{s}$ in the visual input space of the pretrained VLM, while a pixel decoder $g_{\mathrm{pix}}$ reconstructs the input image from the same $z$.

\begin{equation}
\hat{s} = g_{\mathrm{sem}}(z),
\qquad
\hat{x} = g_{\mathrm{pix}}(z).
\end{equation}

\textbf{Training objective.}
Let $s$ denote the visual embeddings fed into the language backbone of the pretrained VLM.
We supervise $z$ with semantic alignment loss
$\mathcal{L}_{\mathrm{sem}}$ and pixel reconstruction loss
$\mathcal{L}_{\mathrm{rec}}$.
The semantic loss $\mathcal{L}_{\mathrm{sem}}$ combines
MSE and cosine-distance losses between $\hat{s}$ and $s$,
while the reconstruction loss is defined as
$\mathcal{L}_{\mathrm{rec}}
= \lVert \hat{x}-x\rVert_1 + \mathrm{LPIPS}(\hat{x},x)$.
The overall tokenizer objective is
\begin{equation}
    \mathcal{L}_{\mathrm{AF}}
    =
    \mathcal{L}_{\mathrm{rec}}
    +
    \lambda_{\mathrm{sem}}\mathcal{L}_{\mathrm{sem}}
    +
    \beta_{\mathrm{KL}}\mathcal{L}_{\mathrm{KL}},
\end{equation}
where $\mathcal{L}_{\mathrm{KL}}$ regularizes the variational
posterior toward a standard Gaussian.
We keep the pretrained vision encoder frozen and optimize only
the AF module, variational bottleneck, semantic projector,
and pixel decoder.
After tokenizer training, we further fine-tune the pixel
decoder $g_{\mathrm{pix}}$ with an additional adversarial
loss~\citep{goodfellow2014gan}.

As shown in Tab.~\ref{tab:reconstruction}, AF applied to Qwen3.5-9B~\citep{qwen3.5} achieves reconstruction quality comparable to or better than that of reconstruction-oriented VAEs.
At the same time, AF retains 99.8\% of the pretrained VLM's understanding performance~(Tab.~\ref{tab:understanding}), demonstrating that high-fidelity reconstruction and visual understanding can coexist in a compact latent space ($d_z=256$).
These unified visual token space $z$ also allows generated visual tokens to be directly reused as input context in multi-turn generation systems, avoiding information loss from repeated pixel decoding and vision re-encoding.

\subsection{Conditional Generation with Block-wise MoT}
\label{sec:method-mot}

Given the AF visual tokens $z$, we model conditional image generation through flow matching.
Let $c$ denote the multimodal conditioning input including text and images, processed by the pretrained VLM with parameters $\theta$.
To predict the velocity field for noisy visual latents $z_t$ conditioned on the VLM representations of $c$, we employ Block-wise MoT, a parameter-efficient variant of Mixture-of-Transformers~(MoT;~\citealt{liang2025mixtureoftransformers}).

\begin{figure}[t]
  \centering
  \hfill
  \begin{subfigure}[t]{0.20\textwidth}
    \centering
    \includegraphics[width=\linewidth]{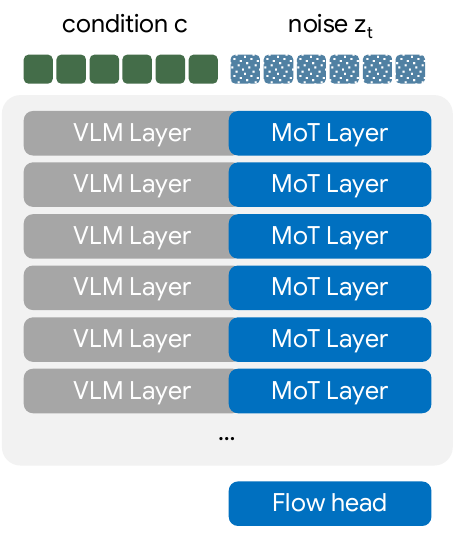}
    \caption{Full MoT}
    \label{fig:mot-a}
  \end{subfigure}\hfill%
  \begin{subfigure}[t]{0.20\textwidth}
    \centering
    \includegraphics[width=\linewidth]{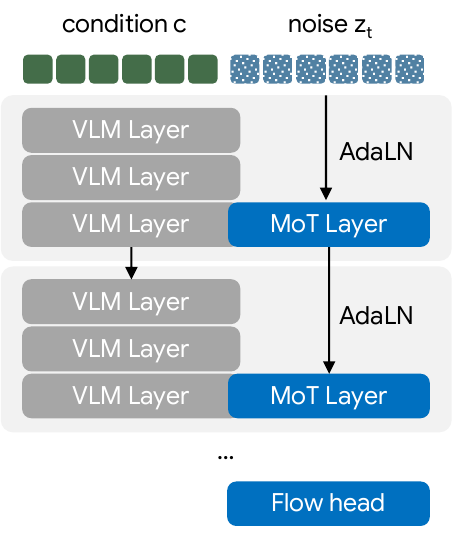}
    \caption{Block-wise MoT}
    \label{fig:mot-b}
  \end{subfigure}\hfill%
  \begin{subfigure}[t]{0.27\textwidth}
    \centering
    \includegraphics[width=\linewidth]{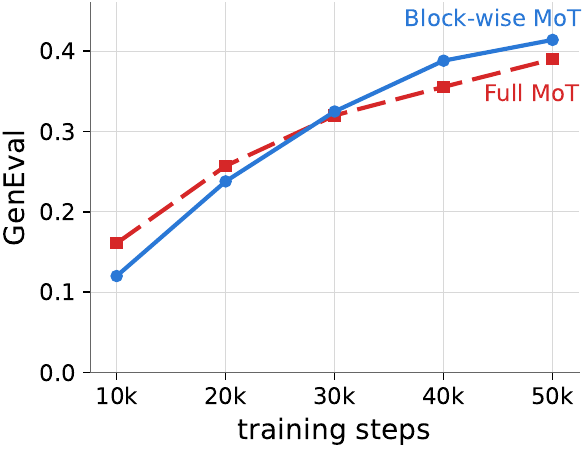}
    \caption{Full vs. Block-wise MoT}
    \label{fig:mot-c}
  \end{subfigure}\hfill%
  \begin{subfigure}[t]{0.27\textwidth}
    \centering
    \includegraphics[width=\linewidth]{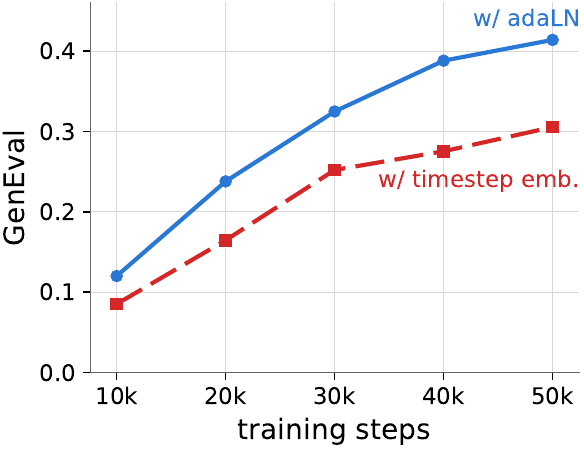}
    \caption{Timestep conditioning}
    \label{fig:mot-d}
  \end{subfigure}\hfill\null
\caption{
\captitle{Block-wise MoT.}
(a) Full MoT pairs every VLM layer with a generation-specific MoT layer.
(b) Block-wise MoT instead groups consecutive VLM layers and performs generative interaction only at block boundaries, with timestep conditioning through adaLN-single.
(c) GenEval during text-to-image pretraining for full MoT and Block-wise MoT using AF visual tokens.
(d) Injecting timestep through adaLN-single versus adding the timestep embedding to the generative stream input.
}
  \label{fig:mot}
\end{figure}

\textbf{Block-wise MoT.} Conventional MoT couples each VLM layer with a generation-specific expert, enabling fine-grained interaction throughout the transformer hierarchy but introducing substantial parameter overhead (Fig.~\ref{fig:mot-a}). 
However, such large generation-specific capacity may be unnecessary if visual generation can share and leverage the VLM's multimodal knowledge through the shared visual tokens $z$.
Motivated by this, we introduce Block-wise MoT, which groups consecutive VLM layers into blocks and performs generative interaction only at block boundaries (Fig.~\ref{fig:mot-b}).
As shown in Fig.~\ref{fig:mot-c}, Block-wise MoT achieves downstream generation comparable to full MoT with only about one-quarter as many generation-specific layers.

Given a noisy visual latent $z_t$, we initialize the generative stream as $g_0=\mathrm{MLP}(z_t)\,\in \R^{N \times d_g}$, where $d_g$ is the width of the generative stream. Let $\ell_1<\cdots<\ell_B$ denote the boundaries of $B$ VLM blocks. The generative stream is updated once at each block boundary:
\begin{equation}
    g_b
    =
    \mathcal{G}_b
    \left(
        g_{b-1},
        u_{\theta}^{(\ell_b)}(c),
        t
    \right),
    \qquad b=1,\ldots,B,
\end{equation}
where $u_{\theta}^{(\ell_b)}(c)$ denotes the attention keys and values directly reused from the VLM at layer $\ell_b$, and $\mathcal{G}_b$ is the corresponding generation block.

Each generation block uses queries from the generative stream to attend to keys and values from both streams.
In this sparse interaction setting, we find that the choice of timestep conditioning has a substantial impact on performance.
As shown in Fig.~\ref{fig:mot-d}, modulating each generation block with adaLN-single~\citep{chen2023pixartalpha} substantially improves performance over simply adding timestep embeddings to the generative stream input, as in conventional MoT.

After the final generation block, $g_B$ provides multimodal conditioning for a flow prediction head, which estimates the velocity field from the noisy latent $z_t$ and timestep $t$:
\begin{equation}
    v_{\phi}(z_t,t,c)
    =
    \mathrm{Pred}_{\phi}(z_t,t,g_B),
\end{equation}
where $g_B$ is the final generative representation conditioned on the multimodal input $c$ through the VLM, and $\phi$ denotes the generation-specific parameters.

We optimize the standard flow matching objective $\mathcal{L}_{\mathrm{FM}}=\E[\|v_\phi(z_t,t,c)-(\epsilon-z)\|^2]$, where $z_t=(1-t)z+t\epsilon$ and $\epsilon\sim\mathcal{N}(0,\mI)$.
For the prediction head, we adopt the wide-head DDT architecture~\citep{wang2026ddt,zheng2026diffusion}.
For timestep sampling, we apply the shifting strategy of \citet{esser2024scaling}.

\section{Experiments}
\label{sec:experiments}

We evaluate the effectiveness of \textit{Integral} as a unified multimodal modeling approach that leverages pretrained VLM knowledge for visual generation.
First, we show that \textit{Integral} achieves strong multimodal understanding, text-to-image generation and image editing performance with only 0.89B generation-specific parameters~(Sec.~\ref{sec:exp-main}).
Moreover, \textit{Integral} achieves strong performance on knowledge-intensive generation tasks, suggesting that pretrained multimodal knowledge can be effectively leveraged for visual generation~(Sec.~\ref{sec:exp-knowledge}).
We further provide controlled comparisons to show that representation sharing enables visual generation to leverage pretrained multimodal knowledge~(Sec.~\ref{sec:exp-analysis}).

\subsection{Setup}
\label{sec:exp-setup}

\begin{table}[t]
\centering
\caption{
\captitle{Multimodal understanding and generation.}
$\dagger$~a
score obtained with rewritten prompts,
and $\S$~a score obtained with chain-of-thought reasoning.
$\ddagger$~Emu3-Chat and Emu3-Gen are reported in one row for convenience.
}
\label{tab:main_results}
\vspace{-5pt}
\setlength{\tabcolsep}{3pt}
\resizebox{\textwidth}{!}{%
\begin{tabular}{lcccccccccccccc}
\toprule
& & \multicolumn{9}{c}{Visual Understanding} & \multicolumn{4}{c}{Image Generation} \\
\cmidrule(lr){3-11}\cmidrule(lr){12-15}
Model & Params & POPE & MME-P & MME-C & MMB & SEED & SEED2+ & MMStar & MMMU & OCRBench & GenEval & GenEval2 & GenEval++ & DPG \\
\midrule
\multicolumn{15}{l}{\textit{Understanding only}} \\
LLaVA-v1.5~\citep{liu2023visual} & 7B & 85.9 & 1511 & -- & 64.3 & 65.8 & -- & 33.1 & 35.7 & 31.8 & -- & -- & -- & -- \\
LLaVA-OV~\citep{li2024llavaonevision} & 7B & -- & 1580 & 418 & 80.8 & 75.4 & 62.2 & 61.7 & 48.8 & 62.2 & -- & -- & -- & -- \\
Qwen2.5-VL~\citep{bai2025qwen25vl} & 7B & -- & -- & -- & 83.5 & -- & 70.5 & -- & 58.6 & 83.7 & -- & -- & -- & -- \\
GPT-5.2~\citep{openai2025gpt52} & -- & -- & -- & -- & 88.2 & -- & -- & 77.1 & 86.7 & 80.7 & -- & -- & -- & -- \\
Gemini-3 Pro~\citep{google2025gemini3pro} & -- & -- & -- & -- & 93.7 & -- & -- & 83.1 & 87.2 & 90.4 & -- & -- & -- & -- \\
Qwen3.5-9B~\citep{qwen3.5} & 9B & 89.6 & 1716 & 685 & 86.3 & 78.9 & 73.1 & 75.9 & 73.2 & 88.1 & -- & -- & -- & -- \\
\midrule
\multicolumn{15}{l}{\textit{Generation only}} \\
SD3-Medium~\citep{esser2024scaling} & 7.7B & -- & -- & -- & -- & -- & -- & -- & -- & -- & 0.74 & -- & 0.296 & 84.08 \\
FLUX.1-dev$^\dagger$~\citep{flux2024} & 17B & -- & -- & -- & -- & -- & -- & -- & -- & -- & 0.82 & -- & 0.314 & 83.84 \\
FLUX-Kontext~\citep{flux-kontext-2025} & -- & -- & -- & -- & -- & -- & -- & -- & -- & -- & -- & -- & 0.343 & -- \\
HiDream-I1-Full~\citep{cai2025hidream} & 31B & -- & -- & -- & -- & -- & -- & -- & -- & -- & 0.83 & -- & -- & 85.89 \\
GPT-4o-Image~\citep{openai2024gpt4ocard} & -- & -- & -- & -- & -- & -- & -- & -- & -- & -- & 0.84 & -- & 0.739 & 85.15 \\
Seedream 3.0~\citep{gao2025seedream} & -- & -- & -- & -- & -- & -- & -- & -- & -- & -- & 0.84 & -- & -- & 88.27 \\
Qwen-Image~\citep{wu2025qwenimagetechnicalreport} & 29B & -- & -- & -- & -- & -- & -- & -- & -- & -- & 0.87 & 33.8 & -- & 88.32 \\
\midrule
\multicolumn{15}{l}{\textit{Unified understanding and generation}} \\
Emu3$^\ddagger$~\citep{wang2024emu3} & 8.5B & 85.2 & 1244 & -- & 58.5 & 68.2 & 44.6 & -- & 31.6 & 68.7 & 0.54 & -- & -- & 80.60 \\
Transfusion~\citep{zhou2025transfusion} & 7.3B & -- & -- & -- & -- & -- & -- & -- & -- & -- & 0.63 & -- & -- & -- \\
Harmon~\citep{wu2025harmonizing} & 2.7B & 87.6 & 1155 & 321 & 65.5 & 67.1 & 41.6 & 35.3 & 38.9 & 11.2 & 0.79 & -- & -- & -- \\
Show-o2~\citep{xie2025showo2} & 8.9B & -- & 1621 & -- & -- & 69.8 & 61.3 & 56.6 & 48.9 & 32.4 & 0.76 & -- & -- & 86.14 \\
UniTok~\citep{ma2025unitok} & 7B & 83.2 & 1448 & -- & -- & -- & -- & -- & -- & -- & 0.59 & -- & -- & -- \\
Janus-Pro~\citep{chen2025janus} & 7B & 87.4 & 1567 & 260 & 79.2 & 72.1 & 56.3 & 48.3 & 41.0 & 59.0 & 0.80 & -- & 0.246 & 84.19 \\
MetaQuery-XL~\citep{pan2025transfer} & 14B & -- & 1685 & -- & 83.5 & 76.9 & -- & -- & 58.6 & -- & 0.80$^\dagger$ & -- & -- & 82.05 \\
BAGEL~\citep{deng2025bagel} & 14B & -- & 1687 & \textbf{701} & 85.0 & 78.5 & 71.9 & -- & 55.3 & 73.3 & 0.82 & 23.1$^\S$ & 0.371 & -- \\
BLIP3-o 8B~\citep{chen2025blip3} & 8B & -- & 1683 & 647 & 83.5 & 77.5 & -- & -- & 50.6 & -- & 0.84 & -- & -- & 81.60 \\
Ming-UniVision~\citep{huang2025ming} & 18B & -- & -- & -- & -- & -- & 56.8 & -- & 40.3 & 72.4 & 0.85 & -- & -- & 82.12 \\
UniAR~\citep{peng2026uniar} & 11B & -- & -- & -- & -- & -- & -- & -- & 44.3 & 83.3 & 0.85 & -- & -- & -- \\
HYDRA~\citep{qiu2026hydra} & 7.8B & -- & -- & -- & 82.4 & 75.5 & -- & 62.3 & 49.4 & 57.7 & 0.86$^\dagger$ & -- & -- & 86.40$^\dagger$ \\
Tuna-2~\citep{liu2026tuna2} & 12B & -- & -- & -- & -- & -- & 61.1 & -- & 50.7 & 79.7 & 0.87 & -- & -- & \textbf{86.54} \\
\midrule
\rowcolor{gray!20}
\textbf{\textit{Integral} (ours)} & 9.8B & \textbf{89.6} & \textbf{1716} & 695 & \textbf{85.4} & \textbf{78.8} & \textbf{73.2} & \textbf{76.0} & \textbf{73.5} & \textbf{87.3} & \textbf{0.88} & \textbf{39.56}$^\S$ & \textbf{0.743} & 84.76 \\
\bottomrule
\end{tabular}}
\end{table}

\begin{table}[t]
\vspace{-8pt}
\centering
\caption{
\captitle{Instruction-based image editing on ImgEdit-Bench.}
We use GPT-4.1 for evaluation.
}
\vspace{-5pt}
\label{tab:imgedit}
\small
\setlength{\tabcolsep}{4pt}
\resizebox{\textwidth}{!}{%
\begin{tabular}{lcccccccccc}
\toprule
Model & Add & Adjust & Extract & Replace & Remove & Background & Style & Hybrid & Action & Overall$\uparrow$ \\
\midrule
GPT-4o~\citep{openai2024gpt4ocard} & 4.61 & 4.33 & 2.90 & 4.35 & 3.66 & 4.57 & 4.93 & 3.96 & 4.89 & 4.20 \\
\midrule
Instruct-Pix2Pix~\citep{brooks2022instructpix2pix} & 2.45 & 1.83 & 1.41 & 2.01 & 1.44 & 1.44 & 3.55 & 1.20 & 1.46 & 1.88 \\
MagicBrush~\citep{zhang2023magicbrush} & 2.84 & 1.58 & 1.51 & 1.97 & 1.58 & 1.75 & 2.38 & 1.62 & 1.22 & 1.90 \\
AnyEdit~\citep{yu2024anyedit} & 3.18 & 2.95 & 1.14 & 2.49 & 2.21 & 2.88 & 3.82 & 1.56 & 2.65 & 2.45 \\
UltraEdit~\citep{zhao2024ultraedit} & 3.44 & 2.81 & 2.00 & 2.96 & 2.45 & 2.83 & 3.76 & 1.91 & 2.98 & 2.70 \\
ICEdit~\citep{zhang2025icedit} & 3.58 & 3.39 & 1.73 & 3.15 & 2.93 & 3.08 & 3.84 & 2.04 & 3.68 & 3.05 \\
Step1X-Edit~\citep{liu2025step1xedit} & 3.88 & 3.14 & 1.76 & 3.40 & 2.41 & 3.16 & 4.63 & 2.64 & 2.52 & 3.06 \\
BAGEL~\citep{deng2025bagel} & 3.56 & 3.31 & 1.88 & 2.62 & 2.88 & 3.44 & 4.49 & 2.38 & 4.17 & 3.20 \\
UniWorld-V1~\citep{lin2025uniworld} & 3.82 & 3.64 & 2.27 & 3.47 & 3.24 & 2.99 & 4.21 & 2.96 & 2.74 & 3.26 \\
OmniGen2~\citep{wu2025omnigen2} & 3.74 & 3.54 & 1.77 & 3.21 & 2.77 & 3.57 & 4.81 & 2.30 & 4.14 & 3.43 \\
\midrule
\rowcolor{gray!20}
\textbf{\textit{Integral} (ours)} & 3.96 & 3.22 & 2.16 & 4.33 & 3.84 & 3.53 & 4.60 & 2.42 & 3.41 & \textbf{3.50} \\
\bottomrule
\end{tabular}}
\vspace{-6pt}
\end{table}

We instantiate \textit{Integral} with Qwen3.5-9B~\citep{qwen3.5}, keeping the pretrained VLM frozen throughout training.
Training consists of two phases: visual tokenizer training and conditional visual generation training.
For visual tokenization, we first train AF at $256\times256$ resolution, followed by dynamic-resolution training up to $512\times512$.
We then fine-tune the pixel decoder with adversarial supervision to improve perceptual reconstruction quality while keeping the learned visual token space fixed.
For visual generation, we train Block-wise MoT on the learned AF visual tokens in three stages.
We begin with text-to-image pretraining at $256\times256$ resolution, followed by higher-resolution interleaved pretraining to support image-conditioned generation.
Finally, we perform supervised fine-tuning on diverse generation tasks, including text-to-image generation, image editing, multi-reference generation, and interleaved reasoning.
Full training details, including datasets and hyperparameters, are provided in Appendix~\ref{app:training}.

\subsection{Multimodal Understanding and Generation}
\label{sec:exp-main}

We first evaluate the effectiveness of \textit{Integral} as a unified multimodal modeling approach by comparing its multimodal understanding, text-to-image generation, and image editing performance with existing UMMs.

\textbf{Evaluation.}
Our evaluation covers a diverse set of benchmarks for multimodal understanding and generation.
For multimodal understanding, we use POPE~\citep{li2023pope}, MME-P and MME-C~\citep{fu2023mme}, MMBench~\citep{liu2024mmbench}, SEED-Bench~\citep{li2023seed}, SEED-Bench-2-Plus~\citep{li2024seed2plus}, MMStar~\citep{chen2024mmstar}, MMMU~\citep{yue2023mmmu}, and OCRBench~\citep{liu2024ocrbench}.
For text-to-image generation, we use GenEval~\citep{ghosh2023geneval}, DPG-Bench~\citep{hu2024ella}, GenEval2~\citep{kamath2025geneval}, and GenEval++~\citep{ye2025echo}.
To assess multimodal conditional generation, we further evaluate instruction-based image editing on ImgEdit-Bench~\citep{ye2025imgedit}.

\textbf{Baselines.}
We compare \textit{Integral} with representative approaches to unified multimodal modeling across diverse architectural paradigms, including discrete-token autoregressive models (Emu3, UniTok;~\citealp{wang2024emu3,ma2025unitok}), decoupled visual encoding (Janus-Pro;~\citealp{wu2024janus,chen2025janus}), combined autoregressive and diffusion modeling (Transfusion, Show-o2;~\citealp{zhou2025transfusion,xie2025showo2}), unified, continuous visual representations (Harmon, Ming-UniVision;~\citealp{wu2025harmonizing,huang2025ming}), Mixture-of-Transformers (BAGEL;~\citealp{deng2025bagel}), and pretrained VLMs with diffusion-based generation (MetaQuery, BLIP3-o;~\citealp{pan2025transfer,chen2025blip3}).
To further assess performance at comparable model scales, we include recent UMMs with approximately 8--12B parameters that report both visual understanding and generation results: UniAR~\citep{peng2026uniar}, HYDRA~\citep{qiu2026hydra}, and TUNA-2~\citep{liu2026tuna2}.

\textbf{Multimodal understanding.}
As shown in Tab.~\ref{tab:main_results}, \textit{Integral} outperforms existing UMMs of comparable or larger scales on most evaluated understanding benchmarks, including larger models such as BAGEL~\citep{deng2025bagel}, Ming-UniVision~\citep{huang2025ming}, and TUNA-2~\citep{liu2026tuna}.
These results highlight the advantage of building upon pretrained VLMs, with \textit{Integral} retaining 99.8\% of the pretrained model's understanding performance.

\textbf{Text-to-image generation.}
Building upon these strong understanding capabilities, \textit{Integral} achieves competitive text-to-image generation performance with only 0.89B generation-specific parameters.
As shown in Tab.~\ref{tab:main_results}, \textit{Integral} achieves a GenEval score of 0.88, outperforming larger UMMs such as BAGEL (0.82) and TUNA-2 (0.87).
Moreover, \textit{Integral} substantially outperforms BAGEL on GenEval2 and GenEval++.
Notably, BAGEL also adopts a MoT architecture but relies on a separate VAE latent space~(FLUX.1-VAE;~\citealp{flux2024}) and approximately 7B generation-specific parameters.
In contrast, \textit{Integral} combines shared visual representations with a lightweight Block-wise MoT, sharing most of its model capacity across understanding and generation.
We further investigate the generative benefits of representation sharing in Sec.~\ref{sec:exp-analysis}.

\textbf{Image editing.}
Beyond text-to-image generation, \textit{Integral} demonstrates strong multimodal conditional generation capabilities on ImgEdit-Bench~\citep{ye2025imgedit}, achieving an overall score of 3.50 and outperforming existing UMMs such as BAGEL (3.20) and OmniGen2 (3.43).
These results show that \textit{Integral} effectively supports generation conditioned on both visual and textual inputs.

Taken together, these results show that \textit{Integral} achieves strong multimodal understanding and visual generation by sharing most of its model capacity across both capabilities.
Fig.~\ref{fig:qual} further illustrates these capabilities with qualitative examples of image generation and editing.

\subsection{Knowledge-Intensive Generation}
\label{sec:exp-knowledge}

\begin{table}[t]
\centering
\caption{
\captitle{Knowledge-intensive text-to-image generation on WISE-Verified.}
Rows marked \emph{+ CoT} explicitly reason before generation.
Bold marks the best overall score among the unified models.
}
\vspace{-5pt}
\label{tab:wise}
\small
\setlength{\tabcolsep}{4pt}
\resizebox{0.9\textwidth}{!}{%
\begin{tabular}{lcccccccc}
\toprule
Model & Params & Culture & Time & Space & Biology & Physics & Chemistry & Overall$\uparrow$ \\
\midrule
\multicolumn{9}{l}{\textit{Closed-source}} \\
NanoBanana-Pro~\citep{google2025nanobananapro}
    & -- & 0.8975 & 0.8167 & 0.9333 & 0.8167 & 0.8667 & 0.8750 & 0.8760 \\
GPT Image 1.5~\citep{openai2025gptimage15}
    & -- & 0.8900 & 0.6917 & 0.8833 & 0.8000 & 0.7583 & 0.7750 & 0.8250 \\
Qwen-Image-2.0~\citep{zhao2026qwen}
    & -- & 0.8219 & 0.6500 & 0.8992 & 0.7917 & 0.8000 & 0.7479 & 0.7954 \\

\midrule
\multicolumn{9}{l}{\textit{Open-source generation-only}} \\
SD-3.5-large~\citep{stabilityai2024sd35large}
    & 14B & 0.4900 & 0.4083 & 0.4417 & 0.3000 & 0.3750 & 0.2083 & 0.4040 \\
Qwen-Image~\citep{wu2025qwenimagetechnicalreport}
    & 29B & 0.6275 & 0.5250 & 0.5583 & 0.3417 & 0.4833 & 0.2500 & 0.5100 \\
HunyuanImage-3.0~\citep{cao2025hunyuanimage}
    & 83B & 0.5250 & 0.3917 & 0.4833 & 0.3083 & 0.4500 & 0.2417 & 0.4350 \\
FLUX.2-dev~\citep{flux-2-2025}
    & 56B & 0.6650 & 0.5667 & 0.6583 & 0.3667 & 0.5250 & 0.3750 & 0.5650 \\

\midrule
\multicolumn{9}{l}{\textit{Unified understanding and generation}} \\
Janus-Pro-7B~\citep{chen2025janus}
    & 7B & 0.3700 & 0.3500 & 0.2833 & 0.2833 & 0.4000 & 0.2333 & 0.3340 \\
BAGEL~\citep{deng2025bagel}
    & 14B & 0.4125 & 0.3500 & 0.3083 & 0.2000 & 0.4417 & 0.2583 & 0.3520 \\
\rowcolor{gray!20}
\textbf{\textit{Integral} (ours)}
    & 9.8B & 0.6217 & 0.5778 & 0.5250 & 0.4778 & 0.6083 & 0.3722 & \textbf{0.5560} \\

\midrule
\multicolumn{9}{l}{\textit{Chain-of-thought reasoning}} \\
BAGEL + CoT
    & 14B & 0.7800 & 0.6333 & 0.5667 & 0.3750 & 0.5500 & 0.5083 & 0.6280 \\
\rowcolor{gray!20}
\textbf{\textit{Integral} (ours)} + CoT
    & 9.8B & 0.8750 & 0.7111 & 0.8056 & 0.7333 & 0.7278 & 0.6694 & \textbf{0.7877} \\

\bottomrule
\end{tabular}}
\vspace{1pt}
\end{table}

\newlength{\panelht}
\setlength{\panelht}{0.2386\textwidth}
\newlength{\fusionht}
\setlength{\fusionht}{0.2761\textwidth}
\newsavebox{\fusionbox}
\newsavebox{\mmditbox}

\begin{figure}[t]
\centering
\begin{minipage}[t]{0.655\textwidth}
\vspace{0pt}
\centering
\begin{subfigure}[t]{0.48\linewidth}
\centering
\includegraphics[height=\panelht]{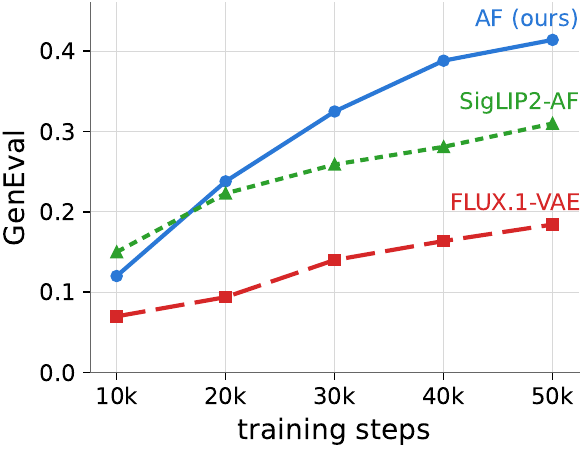}
\caption{Visual tokenizer comparison}
\label{fig:af-variant}
\end{subfigure}\hfill
\begin{subfigure}[t]{0.48\linewidth}
\centering
\includegraphics[height=\panelht]{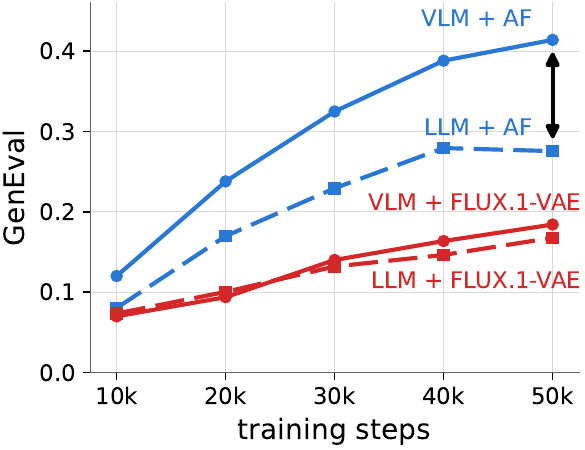}
\caption{VLM vs.\ LLM conditioning}
\label{fig:token-space}
\end{subfigure}
\caption{\captitle{Effect of representation sharing for visual generation.}
(a) Comparison of visual tokenizers under identical VLM conditioning.
(b) Effect of VLM conditioning compared with LLM conditioning, using AF and FLUX.1-VAE.}
\label{fig:rep-sharing}
\end{minipage}\hfill
\begin{minipage}[t]{0.31\textwidth}
\vspace{0pt}
\centering
\begin{subfigure}[t]{\linewidth}
\centering
\includegraphics[height=\panelht]{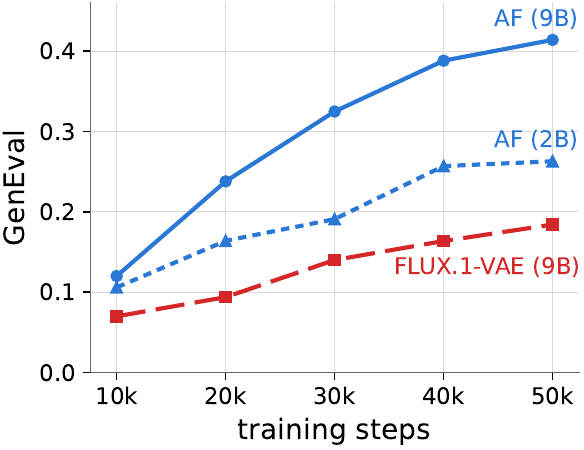}
\caption*{\phantom{(a)}}
\end{subfigure}
\caption{\captitle{Scalability across model sizes.}
Comparison of Qwen3.5-2B and 9B backbones using AF and FLUX.1-VAE.}
\label{fig:scalability}
\end{minipage}
\vspace{-5pt}
\end{figure}

\newsavebox{\kristable}
\sbox{\kristable}{\footnotesize\setlength{\tabcolsep}{3pt}%
\begin{tabular}{lcccc}
\toprule
Model & Fact. & Concept. & Proc. & Overall$\uparrow$ \\
\midrule
GPT-4o~\citep{openai2024gpt4ocard} & 79.80 & 81.37 & 78.32 & 80.09 \\
Gemini 2.0 & 65.26 & 59.65 & 62.90 & 62.41 \\
\midrule
InsPix2Pix~\citep{brooks2022instructpix2pix} & 23.33 & 25.59 & 17.28 & 22.82 \\
OmniGen~\citep{xiao2025omnigen} & 33.11 & 28.02 & 23.89 & 28.85 \\
MagicBrush~\citep{zhang2023magicbrush} & 41.84 & 39.24 & 26.54 & 37.15 \\
AnyEdit~\citep{yu2024anyedit} & 39.26 & 41.88 & 31.74 & 38.55 \\
Emu2 & 45.40 & 37.54 & 34.91 & 39.70 \\
Step1X-Edit~\citep{liu2025step1xedit} & 45.52 & 48.01 & 31.82 & 43.29 \\
BAGEL~\citep{deng2025bagel} & 60.26 & 55.86 & 51.55 & 56.40 \\
\midrule
\rowcolor{gray!20}
\textbf{\textit{Integral} (ours)} & 65.35 & 59.27 & 56.58 & \textbf{60.79} \\
\bottomrule
\end{tabular}}
\setlength{\columnsep}{8pt}
\newlength{\krisintextsep}
\setlength{\krisintextsep}{\intextsep}
\setlength{\intextsep}{0pt}
\begin{wraptable}[13]{r}{\wd\kristable}
\vspace{-0.3\baselineskip}
\centering
\caption{
\captitle{Knowledge-intensive image editing on KRIS-Bench.}
Baseline scores are taken from the official leaderboard. We use GPT-4o for evaluation.
}
\vspace{-8pt}
\label{tab:kris}
\usebox{\kristable}
\end{wraptable}
\AddToHookNext{para/end}{\AddToHookNext{para/end}{\global\intextsep=\krisintextsep}}

Beyond general image generation and editing, we further evaluate \textit{Integral} on knowledge-intensive tasks to examine how effectively it leverages diverse world knowledge encoded in pretrained VLMs for visual generation.

\textbf{Evaluation.}
We evaluate knowledge-intensive text-to-image generation on WISE-Verified, the verified version of WISE-Bench~\citep{niu2026wise}, which assesses generation requiring world knowledge across diverse domains.
For knowledge-intensive image editing, we use KRIS-Bench~\citep{wu2025kris}, which evaluates editing tasks requiring factual, conceptual, and procedural knowledge.

\textbf{Results.}
\textit{Integral} demonstrates strong performance on knowledge-intensive generation and editing tasks, highlighting the benefits of leveraging pretrained multimodal knowledge for visual generation.
As shown in Tab.~\ref{tab:wise}, \textit{Integral} achieves an overall score of 0.556 without explicit chain-of-thought reasoning, consistently outperforming BAGEL (0.352) across all knowledge domains and approaching the much larger FLUX.2-dev (0.565, 56B).
With explicit chain-of-thought reasoning, \textit{Integral} further improves to 0.788, compared with 0.628 for BAGEL under the same setting.
Moreover, as shown in Tab.~\ref{tab:kris}, \textit{Integral} outperforms BAGEL across all three knowledge categories—factual, conceptual, and procedural—achieving a higher overall score (60.79 vs. 56.40) and surpassing specialized image editing models.
These results suggest that \textit{Integral} effectively leverages multimodal knowledge encoded in pretrained VLMs for visual generation.

\subsection{Analysis of Representation Sharing}
\label{sec:exp-analysis}

To investigate the role of representation sharing in leveraging pretrained multimodal knowledge for visual generation, we compare AF with alternative visual tokenizers under controlled settings.
Following our text-to-image pretraining setup, we train each variant for 50K steps with matched generative architectures and training settings.

\textbf{Shared visual representations improve generative learning.}
We first compare AF with FLUX.1-VAE~\citep{flux2024}, a reconstruction-oriented VAE adopted in BAGEL~\citep{deng2025bagel} as a generative latent.
As shown in Figs.~\ref{fig:teaser} and \ref{fig:af-variant}, AF substantially improves generative learning over FLUX.1-VAE, achieving a GenEval score of 0.41 compared with 0.18 under the same training steps (50K).
Given their comparable reconstruction quality (32.2 vs. 31.1 dB PSNR), this gap suggests that sharing visual representations with the pretrained VLM benefits generative learning beyond reconstruction fidelity alone.

\textbf{VLM-native alignment improves generation beyond semantic alignment alone.}
To examine whether the generative benefits of AF arise simply from semantic representations, we compare AF with SigLIP2-AF, which applies the same AF architecture to an external semantic encoder~\citep{tschannen2025siglip}.
As shown in Fig.~\ref{fig:af-variant}, AF outperforms SigLIP2-AF (0.41 vs. 0.31 on GenEval), suggesting that sharing visual representations with the conditioning VLM provides additional generative benefits beyond benefits of semantic space.

\textbf{Shared representations better leverage pretrained VLM conditioning.}
To further examine whether representation sharing facilitates the use of pretrained multimodal knowledge, we compare VLM and text-only LLM conditioning for AF and FLUX.1-VAE.
As shown in Fig.~\ref{fig:token-space}, AF gains substantially more from VLM conditioning than FLUX.1-VAE (+0.13 vs. +0.01 on GenEval).

Taken together, our results highlight the importance of sharing visual representations with the pretrained VLM to effectively leverage its multimodal knowledge for generation.

\subsection{Scalability of Integral}
\label{sec:exp-scalability}

To examine the scalability of \textit{Integral}, we extend our approach to the smaller Qwen3.5-2B backbone.
With this 2B backbone, AF provides a reconstruction PSNR of 32.33 dB while retaining 99.2\% of the pretrained VLM's understanding performance~(Tab.~\ref{tab:understanding-full}).
For visual generation, \textit{Integral} achieves a GenEval score of 0.26 after 50K training steps, outperforming the 9B baseline using FLUX.1-VAE (0.18) under the same training budget.
These results demonstrate that \textit{Integral} can be applied across different VLM scales, preserving pretrained understanding capabilities while enabling visual generation.

\begin{figure}[t]
\centering
\sbox{\fusionbox}{\includegraphics[height=\fusionht]{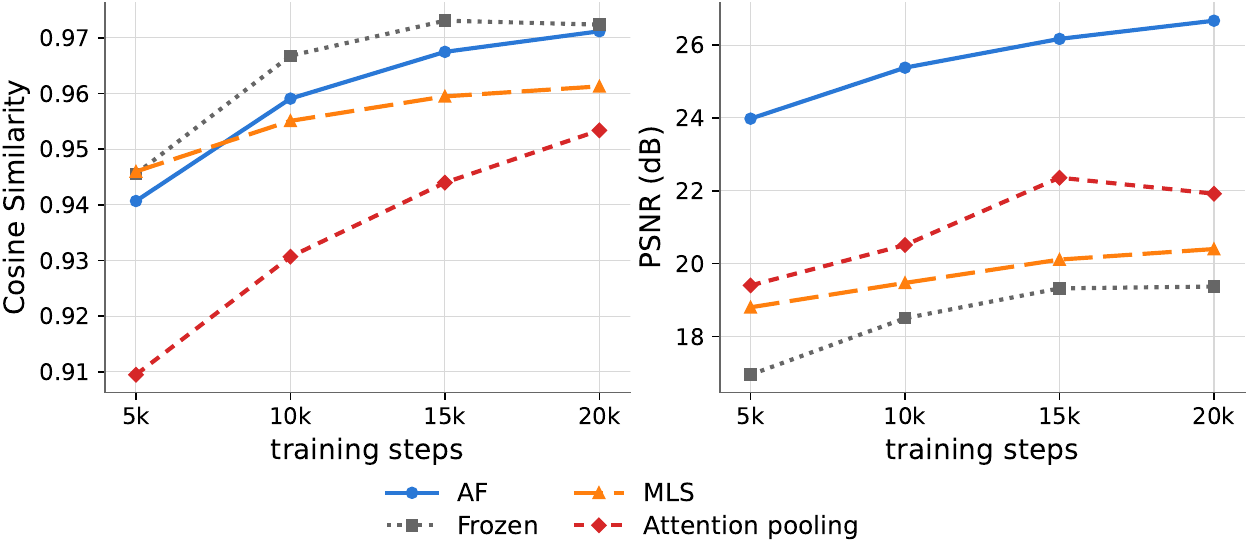}}%
\sbox{\mmditbox}{\includegraphics[height=\panelht]{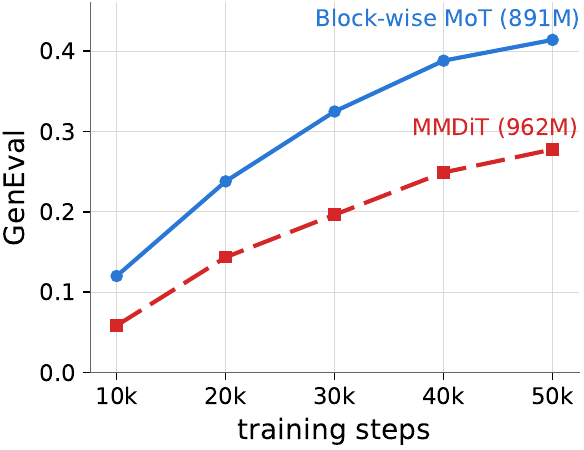}}%
\begin{minipage}[t]{0.63\textwidth}
\vspace{0pt}
\centering
\usebox{\fusionbox}
\caption{\captitle{Comparison of feature fusion methods.}
Reconstruction PSNR and semantic similarity during tokenizer training.}
\label{fig:af-fusion}
\end{minipage}\hfill
\begin{minipage}[t]{0.31\textwidth}
\vspace{0pt}
\centering
\usebox{\mmditbox}\par
\vspace{\dimexpr\ht\fusionbox-\ht\mmditbox\relax}
\caption{\captitle{Block-wise MoT vs.\ MMDiT.}
GenEval performance during training.}
\label{fig:mmdit}
\end{minipage}
\vspace{-5pt}
\end{figure}

\subsection{Component Analysis}
\label{sec:exp-components}

We further evaluate the effectiveness of the two key components of \textit{Integral}, AF and Block-wise MoT, through comparisons with alternative architectures.

\textbf{Abstraction Fusion.}
To evaluate the effectiveness of AF's hierarchical feature fusion, we compare it with multi-layer summation (MLS;~\citealt{singh2026raev2}) and attention pooling under 20K-step tokenizer training.
As shown in Fig.~\ref{fig:af-fusion}, AF substantially improves reconstruction (+6.27 dB PSNR over MLS) while maintaining semantic similarity to the frozen baseline.
In contrast, simpler aggregation methods exhibit a trade-off between reconstruction quality and semantic fidelity, highlighting the effectiveness of AF's fusion architecture.

\textbf{Block-wise MoT.}
To evaluate the effectiveness of Block-wise MoT, we compare it with an MMDiT-based flow head architecture~\citep{xie2025showo2} under the same AF tokenizer, DDT-based predictor, and text-to-image pretraining settings.
As shown in Fig.~\ref{fig:mmdit}, Block-wise MoT achieves higher GenEval scores than MMDiT throughout training, despite using fewer generation-specific parameters (891M vs. 962M).
These results suggest that Block-wise MoT effectively leverages pretrained VLM representations for visual generation with smaller additional capacity.
\section{Related Work}

\textbf{Unified multimodal models.}
Unified multimodal models (UMMs) aim to integrate visual understanding and generation within a single framework.
Early approaches pursued native unification by representing different modalities as discrete tokens and modeling them with a common autoregressive objective~\citep{lu2024unified,chameleon2024chameleon,wang2024emu3}.
Recent work has increasingly incorporated continuous visual representations with diffusion- or flow-based generation objectives alongside autoregressive language modeling~\citep{zhou2025transfusion,xie2025showo2}.
To accommodate the resulting heterogeneity between understanding and generation, several approaches introduce task- or modality-specific pathways while retaining varying degrees of shared computation.
For example, Janus decouples the visual encoding pathways for understanding and generation~\citep{wu2024janus,chen2025janus}, while some approaches introduce specialized transformer experts while retaining shared cross-modal interaction~\citep{liang2025mixtureoftransformers,deng2025bagel,shi2026lmfusion}.
Other approaches couple pretrained visual language models with dedicated generative pathways~\citep{pan2025transfer,shen2025latteflow,lin2025bifrost,pan2026repfusion}, but these approaches rely on separate generative representations.
In contrast, \textit{Integral} constructs a shared visual representation aligned with the pretrained VLM's native features, enabling its existing multimodal knowledge to contribute to image generation.

\textbf{Visual representations for understanding and generation.}
Visual understanding and generation have traditionally relied on different types of visual representations.
Semantic encoders learn transferable, high-level visual features, whereas generative tokenizers are primarily trained to retain information needed for image reconstruction~\citep{radford2021learning,oquab2023dinov2,esser2021taming}.
Recent work has explored pretrained semantic representations for generation~\citep{zheng2026diffusion,singh2026raev2,wang2026decq}, as well as visual tokenizers that combine or separate semantic and reconstructive information~\citep{ma2025unitok,yao2026towards,zhang2025both,yang2026muse,fan2025prism,song2026dualtoken,peng2026uniar}.
These approaches often address the trade-off between semantic preservation and reconstruction quality~\citep{ma2025unitok,yang2026muse,yue2026uniflow}.
In contrast, \textit{Integral} constructs a shared visual representation aligned with the pretrained VLM's native features, enabling its multimodal knowledge to support visual generation.
\section{Conclusion}

We introduced \textit{Integral}, a unified multimodal model that extends a pretrained vision-language model with strong visual generation while preserving multimodal understanding.
Despite using only 0.89B generation-specific parameters, \textit{Integral} achieves strong performance across compositional generation, knowledge-intensive generation, and multimodal image editing.
Our results show that sharing the visual representation space between understanding and generation is key to transferring multimodal knowledge to visual generation.

\section*{Acknowledgments}
We thank Woomin Song and Junhoo Lee for their helpful discussions and feedback.

\bibliography{references}
\bibliographystyle{assets/plainnat}

\newpage
\appendix
\section{Implementation Details}
\label{app:implementation}

\subsection{Architecture}
\label{app:architecture}

\textbf{Abstraction Fusion.}
We use the vision encoder of Qwen3.5-9B~\citep{qwen3.5}, consisting of 27 transformer blocks with patch size 16 and hidden dimension $d_h=1152$.
AF aggregates all 28 representations, including the patch embeddings and the outputs of each transformer block.
Each representation is normalized using fixed per-layer statistics computed from 50K ImageNet training images.
To enable efficient self-attention across encoder layers, we jointly project every four consecutive layers into a single feature, resulting in seven layer-wise features per spatial position.
These features are then aggregated using a self-attention layer with 16 heads, followed by a linear pooler that produces fused representations of dimension $d_f=1152$.

\textbf{Variational bottleneck and decoders.}
The fused representations are compressed into a 256-dimensional latent space through a variational bottleneck, which predicts the mean and log-variance of a diagonal Gaussian distribution.
We sample from this distribution during training and use the posterior mean at inference.
For image reconstruction, we adopt the ViT-XL decoder from RAE~\citep{zheng2026diffusion}, consisting of 28 transformer layers with width 1152.
The semantic projector merges each $2\times2$ neighborhood of visual tokens and projects them to the pretrained VLM's visual input dimension $d_s=4096$.

\textbf{Block-wise MoT.}
We attach eight generation blocks to the 32-layer language backbone of Qwen3.5-9B at intervals of four layers.
To reduce generation-specific parameters, we match the attention QKV dimensions to those of the pretrained VLM while using a reduced hidden dimension of $d_g=2048$ for the remaining computations.
The generative stream is initialized through a two-layer MLP.
Each generation block uses queries from the generative stream and keys and values from both the generative stream and the corresponding frozen VLM layer.
We apply timestep conditioning through adaLN-single~\citep{chen2023pixartalpha} and use a DDT-based prediction head~\citep{wang2026ddt} with two transformer blocks of width 3072.
The noisy visual latents are processed only by the generative pathway, leaving the pretrained VLM unchanged.

\subsection{Training}
\label{app:training}

We train \textit{Integral} in two phases: visual tokenizer training and generation training.
Each phase consists of three stages, as summarized in Tab.~\ref{tab:recipe}.

\textbf{Visual tokenization.}
We first train AF at $256\times256$ resolution, followed by dynamic-resolution training up to $512\times512$.
The tokenizer is optimized with reconstruction, semantic alignment, and KL regularization losses.
The reconstruction loss combines pixel-wise $\ell_1$ and LPIPS losses, while the semantic alignment loss is implemented as
\begin{equation}
    \mathcal L_{\mathrm{sem}}
    =
    \frac{1}{1+r}
    \left[
        \mathrm{MSE}(\hat s,s)
        +
        r\,\mathrm{MSE}\bigl(\bar{\hat s},\bar s\bigr)
    \right],
\end{equation}
where $\bar{\cdot}$ denotes per-token RMS normalization.
We set $r=0.3$, $\lambda_{\mathrm{sem}}=5.0$, and $\beta_{\mathrm{KL}}=10^{-6}$.

In the final tokenizer stage, we fine-tune only the pixel decoder with adversarial supervision while keeping the learned visual latent space fixed.
Following RAE~\citep{zheng2026diffusion}, we use a discriminator based on a frozen DINO ViT-S/8 with trainable heads and optimize the hinge adversarial objective.
The adversarial loss is enabled after 5K steps with a weight of 0.5, scaled by an adaptive weight following VQGAN~\citep{esser2021taming}.

\textbf{Visual generation.}
We use the posterior mean of the trained AF tokenizer as the generation target, standardized using per-channel statistics computed from 50K ImageNet training images.
For flow matching, we adopt logit-normal timestep sampling with a time shift of 4.0~\citep{esser2024scaling}.

Visual generation training consists of three stages.
We first perform text-to-image pretraining at $256\times256$ resolution.
We then conduct interleaved pretraining at resolutions up to $512\times512$ to support multimodal conditional generation.
Finally, interleaved supervised fine-tuning (SFT) covers text-to-image generation, image editing, multi-reference generation, and interleaved reasoning.

\textbf{Datasets.}
For visual tokenizer training, we use DataComp-1B~\citep{datacomp} for the initial training and additionally incorporate BLIP-3o~\citep{chen2025blip3} and RenderedText~\citep{wendler2024renderedtext} for dynamic-resolution training.
The adversarial decoder training uses DataComp-1B.

For visual generation training, text-to-image pretraining uses BLIP-3o, RenderedText, and TextAtlas5M~\citep{wang2025textatlas5m}.
Interleaved pretraining additionally incorporates ScaleEdit-12M~\citep{chen2026scaleedit}, TextSculpt-Data~\citep{lin2026textsculptor}, Pico-Banana-400K, and image reconstruction data from ImageNet-1K~\citep{imagenet15russakovsky} and BLIP-3o.
For interleaved SFT, we use GPT-Image-Edit-1.5M~\citep{wang2025gpt}, Echo-4o-Image~\citep{ye2025echo}, OpenGPT-4o-Image~\citep{chen2025opengpt4oimage}, ScaleEdit-12M, Nano-Consistent-150K, TextSculpt-Data, Zebra-CoT~\citep{li2026zebracot}, and image reconstruction data.

\begin{table}[h]
\centering
\caption{
\captitle{Training recipe for \textit{Integral}.}
Optimization settings, schedules, objectives, and high-level data composition for the three tokenizer-training stages and three generation-training stages.
}
\label{tab:recipe}
\small
\setlength{\tabcolsep}{4pt}
\resizebox{0.85\textwidth}{!}{%
\begin{tabular}{lcccccc}
\toprule
& \multicolumn{3}{c}{Tokenizer} & \multicolumn{3}{c}{Generation} \\
\cmidrule(lr){2-4}\cmidrule(lr){5-7}
& Stage 1 & Stage 2 & Stage 3 & Stage 1 & Stage 2 & Stage 3 \\
\midrule
\multicolumn{7}{l}{\textit{Optimization}} \\
Learning rate & $10^{-4}$ & $10^{-4}\!\rightarrow\!10^{-5}$ & $2\times10^{-5}$ & $10^{-4}$ & $10^{-4}$ & $2\times10^{-5}$ \\
LR scheduler & Constant & Cosine & Constant & Constant & Constant & Constant \\
Warm-up steps & 0 & 500 & 0 & 0 & 1{,}000 & 500 \\
Weight decay & 0.0 & 0.01 & 0.0 & 0.01 & 0.01 & 0.01 \\
Adam $(\beta_1, \beta_2)$ & (0.9, 0.999) & (0.9, 0.999) & (0.9, 0.95) & (0.9, 0.999) & (0.9, 0.999) & (0.9, 0.999) \\
Gradient norm clip & 1.0 & 1.0 & 1.0 & 1.0 & 1.0 & 1.0 \\
EMA decay & -- & -- & -- & 0.9999 & 0.9999 & 0.999 \\
\midrule
\multicolumn{7}{l}{\textit{Schedule}} \\
Training steps & 200K & 400K & 50K & 200K & 50K & 350K \\
GPUs (H100) & 8 & 8 & 8 & 32 & 48 & 48--64 \\
Batch per step & 512 & 262{,}144 tok & 256 & 1024 & 393{,}216 tok & 393{,}216 tok \\
Resolution & 256 & 256--512 & 256 & 256 & $\leq$~512 & $\leq$~512 \\
\midrule
\multicolumn{7}{l}{\textit{Objective}} \\
Flow matching & -- & -- & -- & \checkmark & \checkmark & \checkmark \\
Timestep distribution & -- & -- & -- & \multicolumn{3}{c}{Logit-normal ($0$, $1$), time shift 4.0} \\
Caption dropout & -- & -- & -- & 0.1 & 0.1 & 0.1 \\
\midrule
\multicolumn{7}{l}{\textit{Data sampling ratio}} \\
Text-to-image & -- & -- & -- & 1.00 & 0.400 & 0.400 \\
Image editing & -- & -- & -- & -- & 0.595 & 0.549 \\
Multi-reference & -- & -- & -- & -- & -- & 0.030 \\
Interleaved reasoning & -- & -- & -- & -- & -- & 0.020 \\
Image reconstruction & -- & -- & -- & -- & 0.005 & 0.001 \\
\bottomrule
\end{tabular}}
\end{table}

\section{Additional Evaluation}
\label{app:tokenizer}

We provide additional evaluations of Abstraction Fusion (AF) beyond the main experiments.
We first examine whether AF preserves the pretrained understanding capability across a broader set of benchmarks, and then evaluate its generality across different model scales and vision encoders.

\subsection{Understanding Preservation}
\label{app:understanding}

\begin{table}[t]
\centering
\small
\caption{\captitle{Visual understanding with Abstraction Fusion.}}
\label{tab:understanding-full}
\setlength{\tabcolsep}{3pt}
\resizebox{\textwidth}{!}{%
\begin{tabular}{lccccccccccc}
\toprule
Model & SEED & SEED2+ & MME-P & MME-C & MMMU & POPE & MMB & MMStar & HallusionBench & AI2D & ScienceQA \\
\midrule
Qwen3.5-9B & 78.9 & 73.1 & 1716 & 685 & 73.2 & 89.6 & 86.3 & 75.9 & 72.6 & 87.7 & 94.4 \\
\rowcolor{gray!20}
Qwen3.5-9B w/ AF & 78.8 & 73.2 & 1716 & 695 & 73.5 & 89.6 & 85.4 & 76.0 & 72.5 & 87.8 & 94.6 \\
Qwen3.5-2B & 75.8 & 65.4 & 1585 & 554 & 57.0 & 89.0 & 74.4 & 68.9 & 66.7 & 81.3 & 82.4 \\
\rowcolor{gray!20}
Qwen3.5-2B w/ AF & 75.7 & 65.0 & 1594 & 532 & 55.0 & 88.9 & 73.5 & 66.8 & 66.9 & 80.9 & 81.7 \\
\midrule
Model & RWQA & OCRBench & TextVQA & DocVQA & InfoVQA & MMVP & BLINK & VizWiz & CV-Bench$^{\text{2D}}$ & CV-Bench$^{\text{3D}}$ & Avg. Retention$_{20}$ \\
\midrule
Qwen3.5-9B & 78.2 & 88.1 & 84.3 & 95.6 & 86.9 & 67.3 & 69.3 & 40.0 & 83.0 & 91.1 & 100.0\% \\
\rowcolor{gray!20}
Qwen3.5-9B w/ AF & 78.7 & 87.3 & 83.9 & 95.6 & 86.6 & 67.3 & 67.9 & 39.9 & 83.3 & 90.6 & 99.8\% \\
Qwen3.5-2B & 67.7 & 83.2 & 79.1 & 93.5 & 79.3 & 46.7 & 23.2 & 37.7 & 71.6 & 81.9 & 100.0\% \\
\rowcolor{gray!20}
Qwen3.5-2B w/ AF & 68.8 & 80.8 & 78.9 & 92.0 & 77.3 & 47.3 & 23.4 & 37.1 & 71.9 & 81.3 & 99.2\% \\
\bottomrule
\end{tabular}}
\end{table}

\textbf{Evaluation.} We evaluate on POPE~\citep{li2023pope}, MME-P and MME-C~\citep{fu2023mme}, MMBench~\citep{liu2024mmbench}, SEED-Bench~\citep{li2023seed}, SEED-Bench-2-Plus~\citep{li2024seed2plus}, MMStar~\citep{chen2024mmstar}, MMMU~\citep{yue2023mmmu}, OCRBench~\citep{liu2024ocrbench}, RealWorldQA~\citep{xai2024grok15v}, HallusionBench~\citep{guan2024hallusionbench}, AI2D~\citep{kembhavi2016ai2d}, ScienceQA~\citep{lu2022learn}, TextVQA~\citep{singh2019textvqa}, DocVQA~\citep{mathew2021docvqa}, InfoVQA~\citep{mathew2022infographicvqa}, MMVP~\citep{tong2024eyes}, BLINK~\citep{fu2024blink}, VizWiz~\citep{gurari2018vizwiz}, and CV-Bench~\citep{tong2024cambrian1}.

\textbf{Results.}
As shown in Tab.~\ref{tab:understanding-full}, Qwen3.5-9B with AF retains 99.8\% of the pretrained VLM's understanding performance on average across 20 benchmarks.

\subsection{Generality across Model Scales and Vision Encoders}
\label{app:af-generality}

We apply AF to Qwen3.5-2B and SigLIP2 (\texttt{google/siglip2-so400m-patch16-256};~\citealp{tschannen2025siglip}) to evaluate its applicability across model scales and vision encoders.
For Qwen3.5, we measure understanding retention relative to the pretrained model, while for SigLIP2, we evaluate ImageNet zero-shot classification accuracy.
As shown in Tab.~\ref{tab:af-generality}, AF preserves the original semantic capabilities while achieving reconstruction PSNRs of 32.33 dB and 34.08 dB on Qwen3.5-2B and SigLIP2, respectively.
These results demonstrate that AF can support both semantic preservation and image reconstruction across different model scales and vision encoders.

\begin{table}[t]
\centering
\begin{minipage}[t]{0.625\textwidth}
\centering
\caption{
\captitle{Generality of Abstraction Fusion across model scales and vision encoders.}
Reconstruction is evaluated on ImageNet at $256\times256$.
For Qwen3.5, we report understanding retention relative to the corresponding base model.
For SigLIP2, we report ImageNet zero-shot classification accuracy.
}
\label{tab:af-generality}
\small
\setlength{\tabcolsep}{4pt}
\resizebox{\linewidth}{!}{%
\begin{tabular}{lccc}
\toprule
& \multicolumn{1}{c}{Reconstruction}
& \multicolumn{2}{c}{Semantic Preservation} \\
\cmidrule(lr){2-2}\cmidrule(lr){3-4}
Model
& PSNR $\uparrow$
& Understanding Retention $\uparrow$
& Zero-shot Acc. $\uparrow$ \\
\midrule
Qwen3.5-9B~\citep{qwen3.5}
& --
& 100.0\%
& -- \\
\rowcolor{gray!20}
\quad + AF
& 32.20
& 99.8\%
& -- \\
\midrule
Qwen3.5-2B~\citep{qwen3.5}
& --
& 100.0\%
& -- \\
\rowcolor{gray!20}
\quad + AF
& 32.33
& 99.2\%
& -- \\
\midrule
SigLIP2~\citep{tschannen2025siglip}
& --
& --
& 0.83 \\
\rowcolor{gray!20}
\quad + AF
& 34.08
& --
& 0.83 \\
\bottomrule
\end{tabular}}
\end{minipage}\hfill
\begin{minipage}[t]{0.345\textwidth}
\centering
\caption{\captitle{Component ablations of AF.}
All variants are trained for 20K steps.
}
\label{tab:af-ablation}
\small
\setlength{\tabcolsep}{4pt}
\resizebox{\linewidth}{!}{%
\begin{tabular}{lccc}
\toprule
Variant & PSNR$\uparrow$ & rFID$\downarrow$ & Sem.$\uparrow$ \\
\midrule
\rowcolor{gray!20}
AF (ours) & \textbf{26.67} & \textbf{2.26} & \textbf{0.971} \\
w/o per-layer norm \& proj. & 26.36 & 2.77 & 0.964 \\
w/o self-attention & 26.15 & 2.38 & 0.968 \\
w/ attention pooling & 21.92 & 13.12 & 0.953 \\
\bottomrule
\end{tabular}}

\vspace{8pt}
\caption{\captitle{AF without semantic supervision.}}
\label{tab:af-sem-recon}
\scriptsize
\setlength{\tabcolsep}{4pt}
\begin{tabular}{lcc}
\toprule
Method & PSNR$\uparrow$ & rFID$\downarrow$ \\
\midrule
\rowcolor{gray!20}
AF (ours) & 32.2 & 0.15 \\
AF w/o sem. & \textbf{33.7} & \textbf{0.13} \\
\bottomrule
\end{tabular}
\end{minipage}
\end{table}

\section{Additional Ablation Studies}
\label{app:ablation}

We provide additional analyses of AF and Block-wise MoT, including component ablations, the contributions of encoder layers, and computational efficiency.

\subsection{Abstraction Fusion}
\label{app:ablation-af}

\textbf{Component ablations.}
We examine the contributions of individual components in AF under matched 20K-step tokenizer training settings.
As shown in Tab.~\ref{tab:af-ablation}, replacing per-layer normalization and projection with shared counterparts or removing self-attention degrades reconstruction quality and semantic similarity.
Replacing self-attention and linear pooling with attention pooling further reduces both metrics, supporting the effectiveness of the proposed fusion architecture.

\begin{figure}[t]
  \centering
  \begin{subfigure}[t]{0.32\textwidth}
    \centering
    \includegraphics[width=\linewidth]{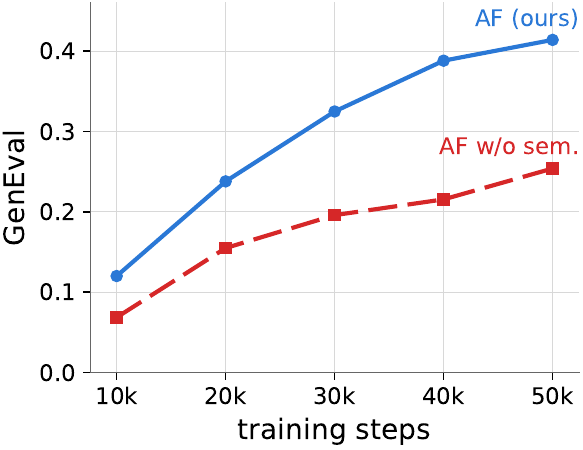}
    \caption{GenEval}
    \label{fig:af-sem-geneval}
  \end{subfigure}\hfill%
  \begin{subfigure}[t]{0.32\textwidth}
    \centering
    \includegraphics[width=\linewidth]{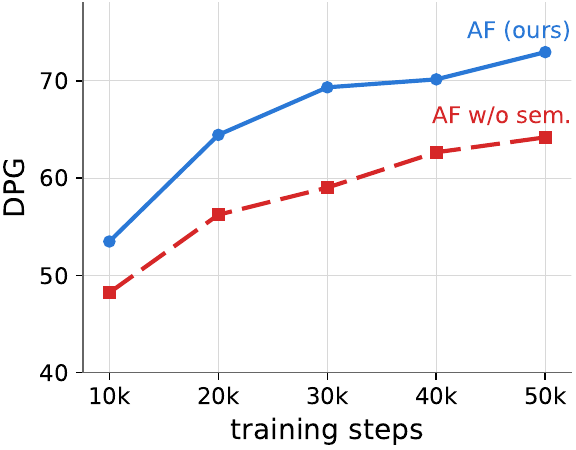}
    \caption{DPG-Bench}
    \label{fig:af-sem-dpg}
  \end{subfigure}\hfill%
  \begin{subfigure}[t]{0.32\textwidth}
    \centering
    \includegraphics[width=\linewidth]{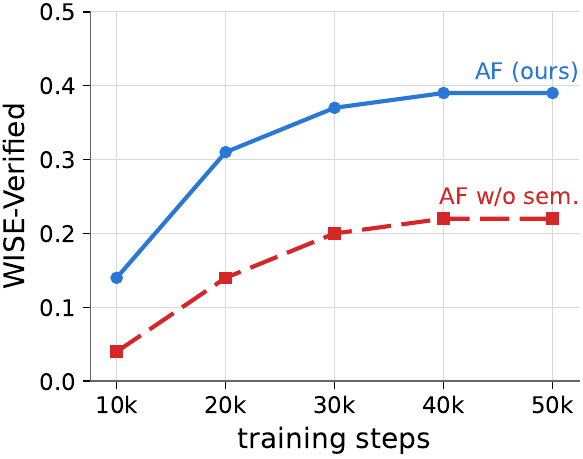}
    \caption{WISE-Verified}
    \label{fig:af-sem-wise}
  \end{subfigure}
  \caption{\captitle{Effect of semantic supervision.}
  Generation performance during 50K-step text-to-image training with and without semantic alignment.}
  \label{fig:af-sem}
\end{figure}

\textbf{Semantic supervision.}
We compare AF with AF w/o sem., which removes the semantic alignment objective while keeping the tokenizer architecture and other training settings unchanged.
As shown in Tab.~\ref{tab:af-sem-recon} and Fig.~\ref{fig:af-sem}, AF w/o sem. achieves higher reconstruction fidelity (33.7 vs.\ 32.2 dB PSNR) but substantially lower generation performance across GenEval, WISE-Verified, and DPG-Bench.
These results further support the importance of semantic supervision for generative learning.

\begin{figure}[t]
  \centering
  \begin{subfigure}[t]{0.32\textwidth}
    \centering
    \includegraphics[width=\linewidth]{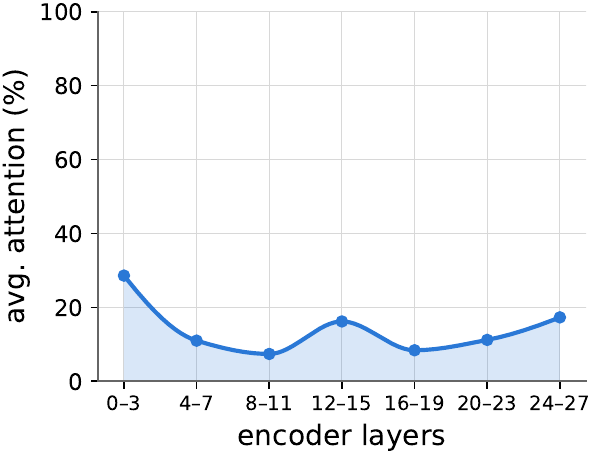}
    \caption{Cross-layer attention}
    \label{fig:af-depth-attention}
  \end{subfigure}\hfill%
  \begin{subfigure}[t]{0.32\textwidth}
    \centering
    \includegraphics[width=\linewidth]{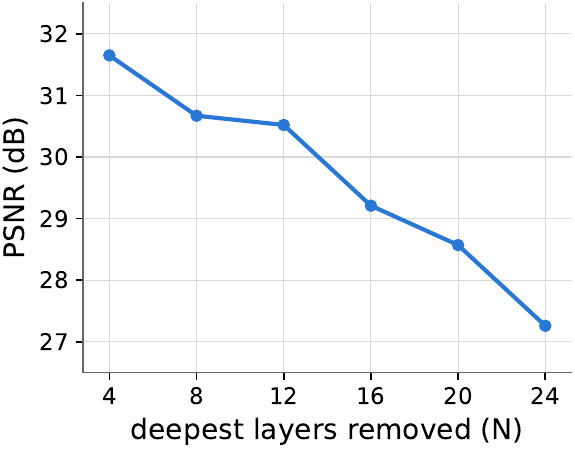}
    \caption{Removing the deepest $N$ layers}
    \label{fig:af-depth-psnr}
  \end{subfigure}\hfill%
  \begin{subfigure}[t]{0.32\textwidth}
    \centering
    \includegraphics[width=\linewidth]{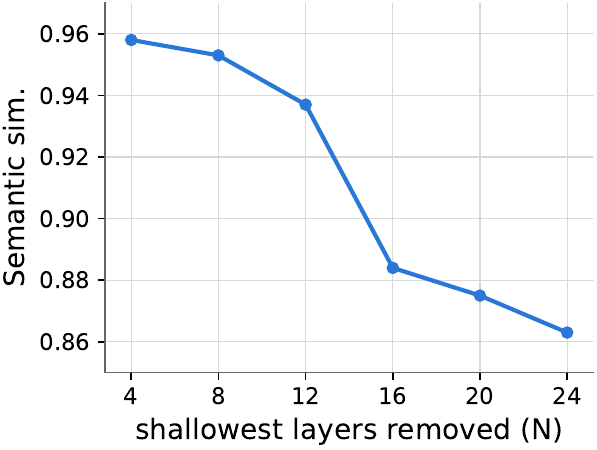}
    \caption{Removing the shallowest $N$ layers}
    \label{fig:af-depth-sem}
  \end{subfigure}
  \caption{\captitle{Contribution of features across encoder depth.}
  (a) Average cross-layer attention across layer groups.
  (b) Reconstruction quality when removing the deepest $N$ layers.
  (c) Semantic similarity when removing the shallowest $N$ layers.}
  \label{fig:af-depth}
\end{figure}

\textbf{Contributions of encoder layers.}
To examine how AF utilizes hierarchical visual features, we analyze the learned attention distribution across encoder layers and progressively remove features from different depths.
As shown in Fig.~\ref{fig:af-depth-attention}, AF distributes attention across the encoder hierarchy, including shallow, intermediate, and deep layers.
Figs.~\ref{fig:af-depth-psnr} and~\ref{fig:af-depth-sem} further show that removing deeper-layer features reduces reconstruction quality, while removing shallower-layer features degrades semantic similarity.
These results indicate that features across different encoder depths contribute to the learned representation.

\subsection{Block-wise MoT}
\label{app:ablation-mot}

We further compare Block-wise MoT with Full MoT, which attaches a generation block to each of the 31 VLM layers.
Both variants use the same AF tokenizer and 50K-step text-to-image pretraining settings.
As shown in Tab.~\ref{tab:mot-ablation}, Block-wise MoT achieves comparable generation performance to Full MoT across GenEval, WISE-Verified, and DPG-Bench while using substantially fewer generation blocks.
These results support the use of sparse interactions with the pretrained VLM for visual generation.

\begin{table}[t]
\centering
\caption{\captitle{Inference overhead of AF.}
Latency and peak GPU memory for a single $800\times800$ image and 100 generated text tokens.
}
\label{tab:af-efficiency}
\small
\setlength{\tabcolsep}{9pt}
\begin{tabular}{lccc}
\toprule
Model & Prefill (ms) & Text generation (ms) & Peak memory (GiB) \\
\midrule
Qwen3.5-9B & 42.6 & 1837 & 17.7 \\
Qwen3.5-9B w/ AF & 54.9 & 1865 & 19.0 \\
\bottomrule
\end{tabular}
\end{table}

\begin{table}[t]
\centering
\caption{\captitle{Block-wise MoT vs.\ Full MoT.}
Generation performance under matched 50K-step training, and denoising latency and computational cost.
}
\label{tab:mot-ablation}
\small
\setlength{\tabcolsep}{5pt}
\resizebox{0.9\textwidth}{!}{%
\begin{tabular}{lccccccc}
\toprule
& & \multicolumn{3}{c}{Generation} & \multicolumn{3}{c}{Inference cost} \\
\cmidrule(lr){3-5}\cmidrule(lr){6-8}
Architecture & Blocks & GenEval$\uparrow$ & WISE-Verified$\uparrow$ & DPG$\uparrow$ & Latency / step$\downarrow$ & Latency / batch$\downarrow$ & Compute$\downarrow$ \\
\midrule
\rowcolor{gray!20}
Block-wise MoT & 8 & 0.41 & 0.39 & 72.95 & 16.1 ms & 858 ms & 387.6 TFLOPs \\
Full MoT & 31 & 0.39 & 0.41 & 72.02 & 52.0 ms & 2659 ms & 970.7 TFLOPs \\
\bottomrule
\end{tabular}}
\end{table}

\subsection{Computational Efficiency}
\label{app:efficiency}

We evaluate the computational overhead of AF for multimodal understanding and the inference efficiency of Block-wise MoT for visual generation.

\textbf{Overhead of AF.}
The AF module and semantic projector replace the original visual processing pathway, increasing its parameter count from 456.0M to 582.1M (+126.1M).
The pixel decoder contains an additional 475.1M parameters and is used for image reconstruction.
We measure inference latency and peak GPU memory using a single $800\times800$ image and 100 generated text tokens.
As shown in Tab.~\ref{tab:af-efficiency}, AF introduces modest overhead in end-to-end understanding inference, increasing latency by 2.1\% and peak memory by 7.3\%.

\textbf{Efficiency of Block-wise MoT.}
We compare the inference costs of Block-wise MoT and Full MoT under the same generation settings.
As shown in Tab.~\ref{tab:mot-ablation}, Block-wise MoT reduces denoising compute by approximately 2.5$\times$ and inference latency by 3.1$\times$ compared with Full MoT.
Together with the generation results in Tab.~\ref{tab:mot-ablation}, these measurements show that Block-wise MoT achieves comparable generation quality at substantially lower computational cost.

\end{document}